\documentclass[authoryear,11pt]{elsarticle}
\usepackage{cancel}
\usepackage{caption}
\usepackage{color}
\usepackage{lscape}
\usepackage{afterpage}
\usepackage{pstricks}
\usepackage{pst-plot}
\usepackage{longtable}
\usepackage{dcolumn}
\usepackage{pst-node}
\usepackage{amsmath}
\usepackage{amssymb}
\usepackage{amsthm}
\usepackage{multirow}
\usepackage{colortab}
\usepackage{color}
\usepackage{rotating}
\usepackage{array}
\usepackage{textcomp}
\usepackage{pst-all}
\usepackage[linesnumbered,ruled,vlined]{algorithm2e}
\usepackage{url}
\usepackage{soul}
\usepackage{hyperref}
\usepackage{xr-hyper}
\theoremstyle{definition} 
\newtheorem{assumption}{Assumption}
\psset{arrows=->, labelsep=3pt, mnode=circle}
\usepackage{lineno}
\usepackage{booktabs}
\usepackage{subcaption}
\usepackage{eurosym}
\usepackage[super]{nth}
\usepackage{csquotes}
\usepackage{bm}
\usepackage{lipsum}
\usepackage{tikz}
\usepackage{acronym}
\usepackage{adjustbox}
\usepackage{amsmath}
\usepackage{amsfonts}
\usepackage{amssymb}
\usepackage{amsthm}
\usepackage{bm}
\usepackage{bbm}
\usepackage{booktabs}
\usepackage{caption}
\usepackage{color}
\usepackage{float}
\usepackage[T1]{fontenc}
\usepackage{makecell}
\usepackage{mathtools}
\usepackage{multirow}
\usepackage{multicol}
\usepackage{setspace}
\usepackage{subcaption}
\usepackage{colortbl}
\usepackage{url}
\usepackage[table]{xcolor} 
\usepackage{tikz}
\usetikzlibrary{shapes.geometric, arrows}
\usepackage{xcolor}
\usepackage{comment}
\usepackage{tabularx}
\newcolumntype{C}{>{\centering\arraybackslash}X}

\newfont{\rams}{msbm10 scaled\magstep1}

\newcommand{\gammadistr}{\mathrm{Gamma}}
\newcommand{\dif}{\mathrm{d}}
\DeclareMathOperator*{\argmin}{argmin}
\DeclareMathOperator*{\argmax}{argmax}

\newtheorem{theorem}{Theorem}[section]

\makeatletter
\newcommand*{\addFileDependency}[1]{%
  \typeout{(#1)}%
  \@addtofilelist{#1}%
  \IfFileExists{#1}{}{\typeout{No file #1.}}%
}
\makeatother

\newcommand*{\myexternaldocument}[1]{%
  \externaldocument{#1}%
  \addFileDependency{#1.tex}%
  \addFileDependency{#1.aux}%
}
\myexternaldocument{suppl_material}

\begin{document}
\pagenumbering{arabic}
\begin{frontmatter}
\title{Bayesian Deck-of-cards-based Ordinal Regression with Sequential Preference Elicitation}

\author[Poz]{Marco Grillo}
\ead{marco.grillo@put.poznan.pl}
\author[Eco]{Silvano Zappalà\corref{cor1}}
\ead{silvano.zappala@phd.unict.it}
\cortext[cor1]{Corresponding author}
\affiliation[Poz]{organization={Institute of Computing Science, Pozna\'{n} University of Technology},
addressline={Piotrowo 2},
postcode={60-965},
city={Pozna\'{n}},
country={Poland}}
\affiliation[Eco]{organization={Department of Economics and Business, University of Catania},
addressline={Corso Italia 55},
postcode={95129},
city={Catania},
country={Italy}}
\begin{abstract}
The Deck-of-cards-based Ordinal Regression (DOR) infers a value function from a ranking of reference alternatives in which the Decision Maker (DM) inserts blank cards between consecutive levels to express preference intensity. DOR, and its stochastic extension (SMAA-DOR), treat these answers as hard constraints defining a set of compatible value functions. We propose B-DOR, a probabilistic reformulation of DOR in which each pair of adjacent levels yields an ordinal observation, the declared direction and the number of cards, modelled through a cumulative-link likelihood that relates the number of blank cards to the latent value difference between alternatives. Two Bayesian inference algorithms are proposed: BAYES-DOR samples the whole posterior distribution by Hamiltonian Monte Carlo; FTRL-DOR tracks the maximum a posteriori estimate by constrained convex optimization. Moreover, through a multi-step elicitation process, elicitation can be spread over several short sessions reducing the cognitive burden on the DM. Both algorithms enjoy logarithmic regret bounds for prediction that hold for any sequence of DM responses and that guide the choice of the prior hyperparameters. A Monte Carlo study over 768 configurations shows that accuracy grows with the number of sessions, that blank cards add significant information over preference directions alone, that both algorithms maintain good performance under inconsistent answers, and that both outperform DOR and SMAA-DOR. An illustrative application to Italian regional healthcare performance demonstrates the practical applicability of the approach for building composite indicators.
\end{abstract}
\end{frontmatter}
\section{Introduction}
Multiple Criteria Decision Aiding (MCDA) supports a Decision Maker (DM) in evaluating a set of alternatives based on several criteria for ranking, scoring, sorting or choice purpose \citep{greco2016multiple}. A relevant instance is the construction of composite indicators, where the overall score of each alternative depends on weights assigned to each criterion and on the chosen aggregation model \citep{greco2019methodological}. The parameters of the preference model can be elicited directly, by questioning the DM about the criteria, or indirectly, such as AHP \citep{saaty1977scaling} and the Best-Worst Method \citep{rezaei2015best,rezaei2026best} where the parameters model are inferred from qualitative pairwise information provided by the DM (for a novel different approach based, instead, on decision rules see \citealt{corrente2026explainable}). The indirect route known as ordinal regression \citep{jacquet1982assessing}, instead, is based on holistic judgements of a few reference alternatives, and it may be less demanding for the DM; Robust Ordinal Regression (ROR) \citep{greco2008ordinal}, later, recognised that many value functions are usually compatible with the same judgements and drew necessary and possible conclusions from the whole compatible set. The Deck of Cards Method (DCM) \citep{simos1990evaluer,figueira2002determining,abastante2022introduction} applies the indirect framework to define the criteria weights or alternatives' score with a simple way to express both the direction and the intensity of preferences: the DM orders the objects at hand and inserts blank cards between consecutive ones, the number of cards expressing the size of the gap. Conceived for criteria weights and widely used to that end \citep{siskos2015elicitation}, DCM was brought into the Ordinal Regression paradigm by \citet{barbati2024deck}: in the Deck-of-cards-based Ordinal Regression (DOR), the DM sorts the reference alternatives into levels of increasing preference and places blank cards between consecutive levels. A linear programming problem is then solved to infer a value function whose values are as proportional as possible to the positions induced by cards. \citet{corrente2025deck} extended DOR to hierarchical criteria and to other scaling procedures, and combined it with ROR and with the Stochastic Multicriteria Acceptability Analysis (SMAA) \citep{lahdelma1998smaa} (hereafter DOR-SMAA), so that the multiplicity of compatible value functions is reflected in the recommendation.

The DOR family shares one modelling decision: the DM's preference information is translated into hard constraints. In other words, they carve a polytope of compatible value functions, which DOR summarises by its most discriminant solution and SMAA-DOR by a uniform sample. The spread of a uniform sample of the polytope induces robustness, but it doesn't directly address preference information of the DM: it neither shrinks because an answer was given with confidence nor widens because two answers disagree. Also, the methods are one-shot: they turn a given block of preference information into a recommendation, but say nothing about how to revise it when further answers are collected, and answers given on different occasions may contradict one another. In fact, DOR elicits ranking preferences over a whole, possibly large, reference set in a single exercise, but the cognitive burden of the DM grows quickly with the size of the set \citep{kivikangas2025effects}; parsimonious methods preserve decision quality while reducing the mental workload \citep{corrente2024better}, and eliciting the intensity of a preference costs only marginally more than the preference itself \citep{haidinger2026much}, so the information in the blank cards comes at a modest price.

Such statements suggest that opting for an elicitation process in which the DM sorts a few alternatives at a time is a promising research direction. We implement such elicitation process in DOR, iterating over several short sessions, and accumulating information. Probabilistic models of the DM's answers are the natural tool for such a framework, and they have been introduced for several MCDA methods: Bayesian versions of the Best-Worst Method \citep{mohammadi2020bayesian}, and, closest to this work, Bayesian ordinal regression for choice and ranking \citep{ru2022bayesian} and for sorting \citep{ru2023probabilistic}, in which the DM's holistic judgements are noisy observations of an additive value function and the recommendation comes from statistical optimization methods. \citet{grillo2025ordinal} then modelled ordinal regression as a sequential prediction problem: the value function is learned online from a stream of pairwise comparisons, by Bayesian updating or by regularised maximum likelihood, and both learners enjoy regret bounds for prediction that hold for every sequence of answers, without any assumption on the DM. While existing probabilistic MCDA models can account for preference intensity, to the best of our knowledge none uses blank cards to elicit such intensity within a probabilistic ordinal regression framework. This paper fills this gap giving DOR a probabilistic model of the blank-card responses and deriving from it two online inference algorithms, under a common framework that we call Bayesian DOR (B-DOR). In particular, our contributions are the following:
\begin{itemize}
    \item \emph{A probabilistic model for deck-of-cards preference information}(Sections~\ref{sec:bdor_data}-\ref{sec:bdor_clm}). It learns how the DM translates preference strength into cards and assigns positive probability to every answer;
    \item \emph{Two inference algorithms with logarithmic regret guarantees} (Sections~\ref{sec:BAYES}-\ref{sec:hyper}). 
    We look at the inference problem as a sequential prediction problem. We prove logarithmic regret bounds for the prediction of DM answers, valid not only for the preference direction but also for its strength (for FTRL-DOR, under a curvature condition on the collected observations). 
    We demonstrate empirically that robust prediction translates to robust Ordinal regression; 
    \item \emph{A few-shot elicitation procedure} (Section~\ref{sec:algorithm}). After each session, a posterior over the model parameters is produced, and becomes the prior of next session. In this way, the DM sorts a few alternatives at a time and the model is refined rather than refitted. The final output is a ranking of all alternatives, along with the most refined posterior distribution of the model parameters;
    \item \emph{A Monte Carlo study on a total of 768 configurations} (Section~\ref{sec:simulation}). It shows that accuracy grows with the number of sessions; the presence of cards significantly improves accuracy over preferences with directions alone; both algorithms maintain good performance in the presence of inconsistent answers; in a single session, where DOR and SMAA-DOR can also be used, both proposed estimators outperform them; and the recorded regret grows logarithmically, as the theory predicts.
    \item \emph{An illustrative example} (Section~\ref{sec:example}). Based on a real-world scenario where 21 Italian regions have to be ranked on the healthcare indicators of the \enquote{Essential Levels of Care}, it shows how the proposed methodologies work.
\end{itemize}

The paper is organised as follows. Section~\ref{sec:background} introduces the notation, the DOR method and the sequential prediction framework. Section~\ref{sec:b_dor} builds the probability model and presents BAYES-DOR and FTRL-DOR with their regret guarantees. Section~\ref{sec:algorithm} describes the few-shot elicitation procedure. Section~\ref{sec:simulation} reports the simulation study. Section~\ref{sec:example} illustrates a didactic case-work. Finally,  Section~\ref{sec:conclusion} concludes. Proofs and complete result tables are collected in the online supplementary material.

\section{Structure and background}
\label{sec:background}
In this section we introduce the basic concepts and notation required to understand the proposed methods.
\subsection{Value function scalar for alternatives}
\label{sec:basic_concept}
We consider a MCDA problem in which a set of alternatives $A = \{a_1, a_2, \ldots\}$ is evaluated on a coherent family of criteria \citep{Roy1996} $G = \{g_1, \ldots, g_m\}$, where $g_j : A \rightarrow \mathbb{R}$ and $g_j(a_i)$ denotes the performance of alternative $a_i$ on criterion $g_j$. In the following, for the sake of simplicity and without loss of generality, let us assume that all criteria have an increasing direction of preference, that is, the greater $g_j(a)$, the better $a$ performs on $g_j$. Let us denote by $A^R \subseteq A$ a set of reference alternatives. It is used to elicit DM's preferences, while $U(a)$ denotes the overall evaluation assigned to each alternative $a \in A$.

To elicit the DM's preferences and build a ranking over $A$, we adopt a preference model in the form of an additive value function:
\begin{equation*}
	U(a)=U\left(g_1(a), \ldots, g_m(a)\right) = \sum_{j=1}^m u_j\left(g_j(a)\right),
\end{equation*}
where each marginal value function $u_j$ captures the DM's preference structure with respect to criterion $g_j$.

Each marginal value function $u_j$ is assumed to be non-decreasing and piecewise linear, defined over $\gamma_j + 1$ characteristic points $c^0_j \leqslant c^1_j \leqslant \ldots \leqslant c^{\gamma_j}_j$, with corresponding values $0 = u_j\left(c^0_j\right) \leqslant \ldots \leqslant u_j\left(c^{\gamma_j}_j\right)$ to be estimated from the DM's preferences. Within each interval $\left[c^{k-1}_j, c^k_j\right]$, the marginal value function
is linear:
\begin{equation*}
	u_j(a)=u_j\left(c_j^{k-1}\right)+
	\frac{g_j(a)-c_j^{k-1}}{c_j^k-c_j^{k-1}}
	\left(u_j\left(c_j^k\right)-u_j\left(c_j^{k-1}\right)\right),
	\quad g_j(a) \in \left[c_j^{k-1}, c_j^k\right].
\end{equation*}
As defined by \citet{grillo2025ordinal}, to handle the value function better from the theoretical viewpoint, we reparametrise $u_j$ using the increments $w_j^k = u_j\left(c_j^k\right) - u_j\left(c_j^{k-1}\right)$, so that:
\begin{equation}
\label{eq:mod_ua}
	u_j(a)=\sum_{i=1}^{k-1} w_j^i+
	\frac{g_j(a) - c_j^{k-1}}{c_j^k-c_j^{k-1}} w_j^k,
	\quad g_j(a) \in\left[c_j^{k-1}, c_j^k\right].
\end{equation}
Using \eqref{eq:mod_ua}, the monotonicity conditions
$u_j\left(c_j^0\right) \leqslant \ldots \leqslant u_j\left(c_j^{\gamma_j}\right)$ reduce simply to $w_j^1, \ldots, w_j^{\gamma_j} \geqslant 0$, which defines a convex feasible region for the parameters. This is a key property for the theoretical guarantees explained in
Section \ref{sec:b_dor}.

To express the value function in compact vector form, for each alternative
$a \in A$ we introduce the feature column vector
$\boldsymbol{v}_j(a)=\left(v_j^1(a), \ldots, v_j^{\gamma_j}(a)\right)$, where:
\begin{equation*}
	v_j^k(a)=\left\{\begin{array}{ll}
		1 & \text{if } g_j(a) \geqslant c_j^k \\[4pt]
		\dfrac{g_j(a)-c_j^{k-1}}{c_j^k-c_j^{k-1}} &
		\text{if } c_j^{k-1} \leqslant g_j(a) \leqslant c_j^k \\[4pt]
		0 & \text{if } g_j(a)<c_j^{k-1}.
	\end{array}\right.
\end{equation*}
Each component $v_j^k(a) \in [0,1]$ measures the degree to which alternative $a$ exceeds the $k$-th characteristic point on criterion $j$. Using the parameter column vector $\boldsymbol{w}_j=\left(w_j^1, \ldots, w_j^{\gamma_j}\right)$, the marginal value can then be written as:
\begin{equation*}
	u_j\left(a\right)=\sum_{i=1}^{\gamma_j} v_j^i(a) w_j^i
	=\boldsymbol{v}_j(a)^{\top} \boldsymbol{w}_j.
\end{equation*}
Concatenating across all criteria as column vectors
$\boldsymbol{v}(a)=\left[\begin{array}{lll}
	\boldsymbol{v}_1(a) & \ldots & \boldsymbol{v}_m(a)\end{array}\right]$
and $\boldsymbol{w}=\left[\begin{array}{lll}
	\boldsymbol{w}_1 & \ldots & \boldsymbol{w}_m\end{array}\right]$, the comprehensive value function takes the linear form:
\begin{equation} \label{eq:utility}
	U(a)=\boldsymbol{v}(a)^{\top} \boldsymbol{w}.
\end{equation}
This representation reduces the estimation of the value function to the estimation of the non-negative parameter vector $\boldsymbol{w} \in \mathbb{R}^N_+$, where $N=\gamma_1 + \cdots + \gamma_m$ is the total number of parameters.
The maximum achievable value under a given $\boldsymbol{w}$ is:
\begin{equation*}
	\bm{w}^\top \bm{1} = \sum_{i=1}^N w_i = \|\bm{w}\|_1,
\end{equation*}
where $\bm{1} = (1,\ldots,1)^\top$ and $\|\cdot\|_1$ denotes the $\ell_1$-norm.
\subsection{The Deck-of-cards-based Ordinal Regression}
\label{sec:dor}
The DOR is a recently introduced methodology that aims to estimate a value function $U$ by aggregating different criteria. In particular, it combines the DM preference information collected by using the DCM and an ordinal regression approach to infer a value function that is as consistent as possible with the DM's preferences. Summarizing, the procedure is based on few steps:
\begin{itemize}
    \item The DM is asked to rank-order alternatives in $A^R$ into sets $L_1,L_2,\ldots,L_s$ so that alternatives in $L_{h+1}$ are preferred to the ones assigned to $L_{h}$ for all $h=1,\ldots,s-1$;
    \item The DM can include a number of \enquote{blank cards} $e_h$ between the sets $L_h$ and $L_{h+1}$ to increase the difference in the global evaluation of the alternatives in the two sets. The number of blank cards is a measure of the intensity of preference of the DM regarding the alternatives ranked in different sets. The greater $e_h$, the more the difference between the global value of alternatives in $L_{h+1}$ and the global value of alternatives in $L_h$;
    \item Each alternative $a\in A^R$ is assigned a position: if $a\in L_1$, $\nu(a)=0$, while for $a\in L_h$ with $h\geqslant 2$, $\nu(a)=\displaystyle\sum_{p=1}^{h-1}(e_p+1)$;
    \item The parameters of the value function are estimated so that the solution deviates as little as possible from the DM's preference information. In particular, it is obtained by following linear programming problem:
    \begin{equation}
        \begin{array}{c}
            \min \overline{\sigma} = \displaystyle\sum_{a\in A^R}\left(\sigma^+(a)+\sigma^-(a)\right),\;\text{subject to,}\\
            \left.
            \begin{array}{l}
                 E^{Model}\\[4mm]
                 U(a)\geqslant U(a^\prime),\;\text{for all}\; a,a^\prime\in A^R\;\text{such that}\; \nu(a)\geqslant\nu(a^\prime)\\[4mm]
                 U(a)-\sigma^+(a)+\sigma^-(a) = k\cdot\nu(a)+\delta\;\text{for all}\;a\in A^R\\[4mm]
                 k\geqslant 0\\[4mm]
                 \sigma^+(a)\geqslant0,\;\sigma^-(a)\geqslant0\;\text{for all}\;a\in A^R
            \end{array}
            \right\}E^{DM}
        \end{array}
    \end{equation}
    where
    \begin{itemize}
        \item $\sigma^+(a)$ and $\sigma^-(a)$ represent over and under estimation of the global value of alternative $a$, while $k$ is the value of a blank card;
        \item $E^{Model}$ is the set of technical constraints related to the considered function $U$. If, for example, it is defined as a normalised weighted sum, it is composed by the following set of constraints:
        \begin{equation*}
        \left.
        \begin{array}{c}
            U(a) = \displaystyle \sum_{g_j\in G}\omega_j\cdot g_j(a)\\[2mm]
            \omega_j\geqslant 0,\;\text{for all}\;g_j\in G\\[2mm]
            \displaystyle\sum_{g_j\in G}\omega_j = 1
        \end{array}
        \right\}E^{Model}_{WS},
        \end{equation*}
        where $\omega_j$ indicates the weight of criterion $g_j$;
        \item $U(a)\geqslant U(a^\prime)$ guarantees the preference order between reference alternatives is respected;
        \item $U(a)-\sigma^+(a)+\sigma^-(a)=k\cdot \nu(a) + \delta$ imposes that the value assigned to alternatives $a\in A^R$ is proportional to the number of blank cards assigned by the DM. Notice that we are using a DOR variant introduced in \citet{corrente2025deck} that avoids asking the DM to place blank cards between $L_1$ and a \enquote{fictitious zero level} $L_0$;
        \item $k\geqslant 0$, $\sigma^+(a)\geqslant 0$ and $\sigma^-(a)\geqslant 0$ impose the non-negativity of the parameters.
    \end{itemize}
\end{itemize} 
The DOR approach, even if it takes into account richer information than classical ordinal regression problem, has some weaknesses. In particular, it does not consider that multiple value functions may correctly represent the DM's preference information. This aspect has been partially assessed by \cite{corrente2025deck}, where DOR is combined with the SMAA. They calculate ranking statistics using parameter vector samples drawn uniformly from the compatible solution space (the space of compatible value functions defined by constraints in $E^{DM}$). Each parameter vector ($s$) defines a compatible value function $U^{(s)}$ that can be used to rank the alternatives in $A$. Consequently, a rank function $rank^{(s)}$ can be defined, assigning to each alternative $a$ its position in the ranking induced  by the compatible value function associated with $s$. We now introduce some of such statistics that will be useful later.
\paragraph{Rank Acceptability Index (RAI)} It is defined as:
\begin{equation}
    \label{eq:RAI}
    \text{RAI}(a_i,k) = \frac{1}{n}\displaystyle\sum_{s=1}^{n}\mathbb{I}\left(rank^{(s)}(a_i)=k\right)
\end{equation}
where $n$ is the sample size and $\mathbb{I}\left(rank^{(s)}(a_i)=k\right)$ is 1 if $a_i$ is ranked at position $k$, and 0 otherwise. It counts the frequency for which an alternative $a_i$ is ranked at the $k$ position.
\paragraph{Pairwise Winning Index (PWI)}
 It is defined as:
\begin{equation}
    \label{eq:PWI}
    \text{PWI}(a_i,a_j) = \frac{1}{n}\displaystyle\sum_{s=1}^n\mathbb{I}\left(U^{(s)}(a_i)\geqslant U^{(s)}(a_j)\right)
\end{equation}
where $\mathbb{I}\left(U^{(s)}(a_i)\geqslant U^{(s)}(a_j)\right)$ is 1 if $U^{(s)}(a_i)\geqslant U^{(s)}(a_j)$, and 0 otherwise. It counts the frequency for which an alternative $a_i$ is ranked not lower than $a_j$.

\subsection{Sequential Probabilistic Framework for Ordinal Regression}
\label{sec:seq_framework}
Here we bridge the ordinal regression problem with the online learning (or sequential prediction) framework \citep{shalev2012online}. This paradigm evaluates learning algorithms sequentially without imposing strict statistical assumptions on the data generating process - such as independence or identically distributed draws - while still being able to provide theoretical guarantees on inference. This distribution-free property renders the framework inherently robust against data corruption, model misspecification, or inconsistent preference judgments from a DM. In this sequential setting, learning proceeds over discrete trials $t = 1, 2, \ldots, T$. At each trial $t$, the algorithm issues a predictive distribution $p_t$, for any data that can possibly be generated by the DM. After that, an observation $Q_t$ is revealed. The algorithm's performance is evaluated via a loss function $\ell(p_t, Q_t)$. The algorithm then updates its internal parameters using the newly acquired information and proceeds to the next trial. The overall predictive accuracy is measured by the cumulative loss over the sequence:
\begin{equation*}\mathcal{L}_T(\mathrm{alg}) = \sum_{t=1}^T \ell(p_t, Q_t).
\end{equation*}
The algorithm's efficiency is evaluated relative to a reference class of prediction models parameterized by a continuous space $\Phi$. The goal is to minimize regret, defined as the difference between the algorithm's cumulative loss and the cumulative loss of the optimal fixed-parameter model chosen in hindsight:
\begin{equation}
\label{eq:regret}\mathcal{R}_T(\mathrm{alg}) = \mathcal{L}_T(\mathrm{alg}) - \inf_{\boldsymbol{\phi} \in \Phi} \mathcal{L}_T(\boldsymbol{\phi}).
\end{equation}
An algorithm is considered a \enquote{no-regret} learner if its average regret $\mathcal{R}_T / T$ converges to zero as $T$ grows, ensuring worst-case performance guarantees regardless of the sequence of observations. 

\subsubsection{Online Inference Algorithms}
To operationalize this sequential assignment over the joint parameters $\boldsymbol{\phi}$, we use a specific class of probabilities $p_{\boldsymbol{\phi}, t} = P(Q_t| \boldsymbol{\phi})$, also referred to as $p_{\boldsymbol{\phi}}$ when the time step is inferred by the context, and a class of loss functions known as log-loss functions 
$$
\ell(p_{\boldsymbol{\phi}}, Q_t) = -\ln P(Q_t| \boldsymbol{\phi}).
$$
Both are adopted in two foundational classes of online learning algorithms. 

\paragraph{Bayesian Inference}
Bayesian strategies are initialized by a prior density
$\pi(\boldsymbol{\phi})$ on the parameter space $\Phi$. After observing $Q_t$,
the algorithm updates its internal state to form a posterior distribution
incorporating the likelihood $P(Q_t\mid\boldsymbol{\phi})$. We define the data
acquired up to time $t$ as
$\mathcal{D}_t=\{Q_1,\ldots,Q_t\}$, so that Bayes' rule gives
\begin{equation*}\displaystyle\pi(\boldsymbol{\phi} | \mathcal{D}_t) = \displaystyle\frac{ P(Q_t | \boldsymbol{\phi})  \pi(\boldsymbol{\phi} | \mathcal{D}_{t-1})}{\displaystyle\int_{\boldsymbol{\phi}^\prime\in\Phi} P(Q_t | \boldsymbol{\phi}^\prime) \pi(\boldsymbol{\phi}^\prime | \mathcal{D}_{t-1})\, \mathrm{d}\boldsymbol{\phi}^\prime}.
\end{equation*}
Predictions for future trials are made by marginalizing over this posterior. The
exact Bayesian cumulative log-loss over the sequence is the negative logarithm
of the marginal likelihood. The construction provides a principled mechanism
for incorporating prior information, but it requires integration over the
continuous parameter space $\Phi$, often motivating computational
approximations such as Monte Carlo sampling.

\paragraph{Follow the Regularized Leader (FTRL)}
As a computationally lighter alternative, FTRL replaces the full distribution
by a deterministic parameter sequence. At each trial $t$, it selects a
parameter that minimises the past cumulative log-loss plus a regularisation term
$\Omega(\boldsymbol{\phi})$:
\begin{equation*}
    \boldsymbol{\phi}_t = \argmin_{\boldsymbol{\phi} \in\Phi} \left\{ \mathcal{L}_{t-1}(\boldsymbol{\phi}) + \Omega(\boldsymbol{\phi}) \right\}.
\end{equation*}
When the regulariser is chosen as the negative logarithm of the prior ($\Omega(\boldsymbol{\phi}) = -\ln \pi(\boldsymbol{\phi})$), FTRL functions as a sequential Maximum A Posteriori (MAP) estimator.

\section{The Bayesian-DOR (B-DOR) methods}
\label{sec:b_dor}
In this section we integrate the DOR method with a probabilistic model and derive two inference algorithms from it. The algorithms can be used in an online way: the DM is questioned over a sequence of elicitation sessions $t=1,\dots,T$. In each session, the DM distributes a small set of alternatives into levels sorted by preference and places blank cards between consecutive levels. Section~\ref{sec:bdor_data} explains how we record this information as a set of \emph{gap observations}. Section~\ref{sec:bdor_clm} builds a probability model for these observations: each provides information relative to the direction and strength of a pairwise comparison between adjacent levels. Sections~\ref{sec:BAYES} and~\ref{sec:FTRL} then instantiate the two online inference paradigms of Section~\ref{sec:seq_framework}: BAYES-DOR, which maintains a full posterior distribution of rankings, and FTRL-DOR, which tracks a single regularised ranking estimate.

\subsection{From a session to gap observations}
\label{sec:bdor_data}
At session $t$ the DM is shown a subset $S_t \subseteq A^R$ of $k_t=|S_t|$ alternatives. Following the DOR protocol recalled in Section~\ref{sec:dor}, the DM sorts them into ordered levels
\begin{equation*}
L_{t,1}\prec L_{t,2}\prec\dots\prec L_{t,l_t},
\end{equation*}
where $l_t$ denotes the number of levels at session $t$. Alternatives placed in the same level are evaluated as indifferent (having the same global value). 
Notice that $l_t=k_t$ when no two alternatives are tied, and $l_t<k_t$ otherwise. The DM then inserts $e_{t,h}$ blank cards between consecutive levels $L_{t,h}$ and $L_{t,h+1}$, for $h=1,\dots,l_t-1$. The cards express how much better the upper level is: zero cards mean \enquote{minimal} difference, while many cards mean \enquote{more} difference. The protocol fixes in advance a maximum number of cards per gap, $e_{\max}$, so every count lies in $\{0,1,\dots,e_{\max}\}$. Differently from the DOR, we do not convert the cards into position values $\nu(\cdot)$: the model below uses the card counts directly.

A session thus produces two things: the sorting itself (which level is above which, and which alternatives share a level) and the $l_t-1$ card counts. We model the card counts, together with the directions they refer to, as random responses. The observation recorded at session $t$ is
\begin{equation}
\label{eq:obs_set}
    Q_t=\bigl( (L_{t,1},\dots,L_{t,l_t}),\ (e_{t,1},\dots,e_{t,l_t-1}) \bigr).
\end{equation}

Each gap between two adjacent levels is associated with one feature vector. For a level $L$, let $\bar{\boldsymbol{v}}(L)= \displaystyle\frac{1}{|L|}\displaystyle\sum_{a\in L}\boldsymbol{v}(a)$ be the average of the feature vectors of its members, and for the $h$-th gap define
\begin{equation}
\label{eq:diff_vector}
    \boldsymbol{x}_{t,h} \;=~\; \bar{\boldsymbol{v}}(L_{t,h+1})-\bar{\boldsymbol{v}}(L_{t,h}) \;\in\;[-1,1]^N .
\end{equation}
When every level contains a single alternative, $\boldsymbol{x}_{t,h}$ is simply the feature difference of two consecutive alternatives. Because the value function \eqref{eq:utility} is linear in the features, the latent quantity
\begin{equation*}
    z_{t,h}=\boldsymbol{x}_{t,h}^{\top}\boldsymbol{w}
\end{equation*}
equals the difference between the \emph{average} values of the two levels. Note also that every component of $\boldsymbol{x}_{t,h}$ lies in $[-1,1]$, so $\|\boldsymbol{x}_{t,h}\|_\infty\leqslant1$ holds by construction.

We denote by $\mathcal{D}_t=\{Q_1,\dots,Q_t\}$ the information accumulated up to session $t$. Since a session with $l_t$ levels has $l_t-1$ gaps, session $t$ contributes $C_t=l_{t}-1$ \emph{gap observations}, and after $T$ sessions the total number of gap observations is $n_T$:
\begin{equation}
\label{eq:nT}
    n_T=\sum_{t=1}^{T}C_t .
\end{equation}
In other words, $n_T$ counts the elementary (direction, cards) atoms of information collected at $T$. For example, a session in which $k_t=5$ alternatives are fully ranked without ties has $l_t=5$ levels and contributes $C_t=4$ observations; if no ties ever occur, $n_T=\displaystyle\sum_{t=1}^T(k_t-1)$. The regret guarantees of Sections~\ref{sec:BAYES} and~\ref{sec:FTRL} are stated in terms of $n_T$. A degenerate session in which the DM places all alternatives in a single level ($l_t=1$) produces no gap observation and simply does not contribute to the likelihood. Subsets may repeat across sessions, so $S_t\cap S_{t'}\neq\emptyset$ is allowed; repeated gap observations can accumulate as additional evidence.

\subsection{A probability model for the cards}
\label{sec:bdor_clm}
\begin{figure}[h!]
\centering
\begin{tikzpicture}[x=1cm,y=1cm]
  \fill[gray!12] (-3.9,-0.05) rectangle (0,0.78);
  \draw[->] (-4.0,0) -- (8.0,0) node[below right=-2pt] {$z$};
  \draw (0,0.10) -- (0,-0.10) node[below=2pt] {$0$};
  \draw (1.8,0.10) -- (1.8,-0.10) node[below=2pt] {$\theta_1$};
  \draw (3.6,0.10) -- (3.6,-0.10) node[below=2pt] {$\theta_2$};
  \draw (5.8,0.10) -- (5.8,-0.10) node[below=2pt] {$\theta_{e_{\max}}$};
  \node[below=6pt] at (4.7,0) {$\cdots$};
  \node[align=center] at (-2.0,0.42) {\small reversal:\\[-2pt]\small lower level preferred};
  \node at (0.9,0.42) {\small $0$ cards};
  \node at (2.7,0.42) {\small $1$ card};
  \node at (4.7,0.42) {\small $\cdots$};
  \node at (7.0,0.42) {\small $e_{\max}$ cards};
\end{tikzpicture}
\caption{Latent preference scale $z$ partitioned into card bands by thresholds $\theta_m$.}
\label{fig:latent_scale}
\end{figure}
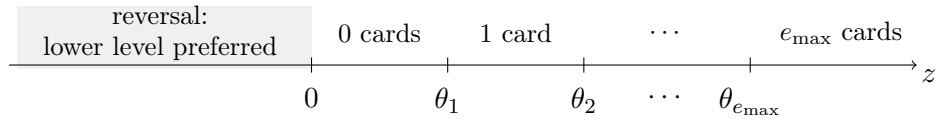

The model rests on one simple picture, sketched in Figure~\ref{fig:latent_scale}. Each gap has a latent \emph{strength} $z=\boldsymbol{x}^{\top}\boldsymbol{w}$, that depends on the difference between the feature vector of the upper level and the feature vector of the lower level, and the parameters $\boldsymbol{w}$ that define the DM response to criteria. This number can fall anywhere on the real line, and its position has a direct meaning. If $z>0$, the upper level is indeed the better one, and the larger $z$, the stronger the preference. If $z=0$, the two levels are equally good. If $z<0$, the lower level is in fact the better one, so the direction declared by the DM would be a mistake.

To connect this scale with the cards, we divide the positive half-line into $e_{\max}+1$ consecutive intervals (one for each possible number of cards) into thresholds $\theta_m$:
\begin{equation}
\label{eq:thresholds}
    0\equiv\theta_0<\theta_1<\dots<\theta_{e_{\max}}<\theta_{e_{\max}+1}\equiv+\infty,
    \qquad
    \theta_m=\sum_{i=1}^{m}\Delta_i,
\end{equation}
parameterised by increments $\Delta_i\geqslant\rho>0$; the floor $\rho$ prevents two thresholds from collapsing onto each other. The interval (or \emph{band}) $[\theta_e,\theta_{e+1})$ contains the continuous values of strength that the DM expresses with exactly $e$ cards. It is clear at this point that $e_{max}$ also defines the dimensionality of a vector of increments $\boldsymbol{\Delta}$. Note that the fact that thresholds can have different values allows for robust fitting of any response function used by the DM that maps strength to a number of cards, and not just a linear trivial one. We collect parameters of the model into a joint vector $\boldsymbol{\phi}=(\boldsymbol{w},\boldsymbol{\Delta})$, with $\boldsymbol{\Delta}=(\Delta_1,\dots,\Delta_{e_{\max}})$.

The DM does not know $z$ exactly: we expect preferences and their strength to be noisy. We model the perceived strength as $z$ plus a standard logistic error. The probability that the DM confirms the direction of the gap \emph{and} places $e$ cards is the probability that the perceived strength falls in the $e$-th band:
\begin{equation}
\label{eq:clm_prob}
    P(\succ,e \mid \boldsymbol{x}, \boldsymbol{\phi})
    \;=\;
    \sigma\!\left(z-\theta_{e}\right)-\sigma\!\left(z-\theta_{e+1}\right),
    \qquad e\in\{0,1,\dots,e_{\max}\},
\end{equation}
where $\sigma(s)=(1+e^{-s})^{-1}$ is the logistic function. Suppose we observe $e<e_{\max}$, the probability in \eqref{eq:clm_prob} is maximized when the score lies at the centre of the observed card band:
$$
z^{*}=\frac{\theta_e+\theta_{e+1}}{2},
\qquad
\max_z P(\succ,e\mid\boldsymbol{x},\boldsymbol{\phi})
=\tanh\left(\frac{\theta_{e+1}-\theta_e}{4}\right)
=\tanh\left(\frac{\Delta_{e+1}}{4}\right).
$$
Notice that multiplying both $z$ and $\Delta_{e+1}$ by $c>1$ raises the maximum to $\tanh(c\Delta_{e+1}/4)$, so a maximum likelihood estimator would push some components of both $\boldsymbol{w}$ and $\boldsymbol{\Delta}$ simultaneously to infinity. This is why our, soon to be described, algorithms use prior distributions as regulariser.

For a fixed orientation of a pair, the opposite direction is represented by
reversing the feature vector. Therefore,
$$
\sum_{e=0}^{e_{\max}}
\left[
P(\succ,e\mid\boldsymbol{x},\boldsymbol{\phi})
+
P(\succ,e\mid-\boldsymbol{x},\boldsymbol{\phi})
\right]
=
\sigma(z)+\sigma(-z)=1.
$$
Thus, each term in \eqref{eq:clm_prob} is a proper probability for one
direction-and-card response. It is a cumulative-link (ordered logit) term
\citep{mccullagh1980regression}; when $e_{\max}=0$, it reduces to the binary
logistic term used by \citet{grillo2025ordinal}.
We combine gap observations through the adjacent-gap composite likelihood
\begin{equation}
\label{eq:step_likelihood}
    P(Q_t \mid \boldsymbol{\phi}) \;=\; \prod_{h=1}^{l_t-1} P(\succ,e_{t,h} \mid \boldsymbol{x}_{t,h}, \boldsymbol{\phi}),
\end{equation}
with corresponding log-loss
\begin{equation}
\label{eq:logloss}
    \ell(p_{\boldsymbol{\phi}}, Q_t) \;=\; -\sum_{h=1}^{l_t-1} \ln P(\succ,e_{t,h} \mid \boldsymbol{x}_{t,h}, \boldsymbol{\phi}),
\end{equation}

\subsection{BAYES-DOR}
\label{sec:BAYES}
The first inference strategy is fully Bayesian: instead of committing to a single value function, it maintains a distribution over $\boldsymbol{\phi}$, which records how much uncertainty is left after the sessions seen so far. The distribution starts from a prior, which also regularises the inference when data are scarce. We assign independent Gamma priors to the weights and shifted-Gamma priors to the threshold increments,
\begin{equation}
\label{eq:prior}
    \pi(\boldsymbol{w},\boldsymbol{\Delta})
    =\prod_{i=1}^N \gammadistr(w_i\mid\alpha,\beta)
     \prod_{m=1}^{e_{\max}}
     \gammadistr(\Delta_m-\rho\mid\alpha_{\Delta},\beta_{\Delta}),
     \qquad \Delta_m\geqslant\rho .
\end{equation}
The shape parameters are restricted to $\alpha,\alpha_{\Delta}\geqslant1$, so to keep each marginal prior log-concave on its support.

Given the data $\mathcal{D}_T$, the likelihood and the posterior are
\begin{equation}
\label{eq:likelihood}
    P(\mathcal{D}_T \mid \boldsymbol{\phi}) = \prod_{t=1}^T P(Q_t \mid \boldsymbol{\phi}),
    \qquad\qquad
    \pi(\boldsymbol{\phi} \mid \mathcal{D}_T) \propto P(\mathcal{D}_T \mid \boldsymbol{\phi})\, \pi(\boldsymbol{\phi}),
\end{equation}
and the posterior can be computed one session at a time:
\begin{equation}
\label{eq:posterior_recursive}
    \pi(\boldsymbol{\phi} \mid \mathcal{D}_t) \propto P(Q_t \mid \boldsymbol{\phi})\, \pi(\boldsymbol{\phi} \mid \mathcal{D}_{t-1}), \qquad \pi(\boldsymbol{\phi} \mid \mathcal{D}_0) \equiv \pi(\boldsymbol{\phi}).
\end{equation}
In words, the posterior after one session becomes the prior for the next one. This recursion is the engine of the multi-session procedure of Section~\ref{sec:algorithm}. Each elementary loss $-\ln P(\succ,e\mid\boldsymbol{x},\boldsymbol{\phi})$ is convex in $\boldsymbol{\phi}$ (Lemma~\ref{lem:appA_convexity} in the online supplementary material; see also \citealp{pratt1981concavity,burridge1981note}), and the priors are log-concave, so the negative log-posterior is a convex function on its domain.

\paragraph{Regret guarantee}
The cumulative loss of BAYES-DOR is the log-loss of its \emph{exact} posterior predictive distributions,
\begin{equation}
\label{eq:bdor_bayesian_cumulative_loss}
    \mathcal{L}_T(\mathrm{BAYES}\text{-}\mathrm{DOR})
    =-\sum_{t=1}^{T}\log\int P(Q_t\mid\boldsymbol{\phi})\,
      \pi(\boldsymbol{\phi}\mid\mathcal D_{t-1})\,\dif\boldsymbol{\phi},
\end{equation}
which equals the negative log marginal likelihood of the whole sequence. We compare it with the best fixed parameter, chosen in hindsight, in the compact class
\begin{equation}
\label{eq:compact_class_maintext}
    \Phi_{\rho,M}
    =\Bigl\{(\boldsymbol{w},\boldsymbol{\Delta})
      \in[0,\infty)^N\times[\rho,\infty)^{e_{\max}}
      :\ \|\boldsymbol{w}\|_1+\|\boldsymbol{\Delta}\|_1\leqslant M\Bigr\},
    \qquad M>e_{\max}\rho .
\end{equation}
Here $M$ is a fixed finite radius that bounds the joint scale of the value-function weights and threshold increments. The condition $M>e_{\max}\rho$ ensures that $\Phi_{\rho,M}$ is non-empty. Its role is to provide a compact comparator class over which the constants in the regret bound are uniform.

\begin{theorem}[Uniform logarithmic regret for BAYES-DOR]
\label{thm:bayes_dor_regret}
Assume the shape condition $\alpha,\alpha_{\Delta}>1$, and recall that the gap features satisfy $\|\boldsymbol{x}_{t,h}\|_\infty\le1$ by construction. Let BAYES-DOR use the prior \eqref{eq:prior}. Then there exists a finite constant $C^{\mathrm B}_{\rho,M}$, independent of $T$ and of the realised data sequence, such that for every $T$ with $n_T\geqslant1$,
\begin{equation}
\label{eq:bayes_regret_maintext}
    \mathcal{L}_T(\mathrm{BAYES}\text{-}\mathrm{DOR})
    -\inf_{\boldsymbol{\phi}\in\Phi_{\rho,M}}
      \mathcal{L}_T(\boldsymbol{\phi})
    \ \leqslant\ C^{\mathrm B}_{\rho,M}+A_{\rho}\log n_T,
    \qquad
    A_{\rho}=\alpha N+\alpha_{\Delta}\,e_{\max}.
\end{equation}
\end{theorem}

\noindent
The proof, together with the explicit form of $C^{\mathrm B}_{\rho,M}$, is given in the online supplementary material (Theorem~\ref{thm:appA_bayes}); it relies on the classical mixture/prior-mass argument for Bayesian prediction under logarithmic loss \citep{kakade2004online,cesabianchi2006prediction}.

Theorem~\ref{thm:bayes_dor_regret} says that BAYES-DOR is a no-regret predictive procedure: its cumulative log-loss exceeds that of the best fixed parameter in $\Phi_{\rho,M}$ only by a term that grows logarithmically in the number $n_T$ of gap observations, so the average gap between the two vanishes as $n_T$ grows. In practical terms: after sufficiently many sessions, BAYES-DOR predicts the DM's next answers (directions and card counts) essentially as well as the best single value function that could have been chosen with hindsight, and this holds for \emph{every} realised sequence, even if the DM answers inconsistently: no assumption is made on how the DM generates the responses. The constants are not optimised; refined arguments would improve $A_\rho$, but not the logarithmic rate.

In practice the posterior is not available in closed form, and our implementation approximates it by Hamiltonian Monte Carlo sampling \citep{neal2011mcmc,hoffman2014nuts}. The sampled parameters are used to compute posterior summaries of the recommendation, such as the RAI \eqref{eq:RAI} and the PWI \eqref{eq:PWI}.

\subsection{FTRL-DOR}
\label{sec:FTRL}
Sampling the posterior after every session can be computationally demanding. FTRL-DOR is the lighter alternative: it replaces the distribution by a single point estimate, following the Follow-The-Regularised-Leader principle of Section~\ref{sec:seq_framework}. At session $t$ it selects the parameter that best explains all past sessions, regularised by the prior:
\begin{equation}
\label{eq:ftrl_map}
    \boldsymbol{\phi}_t^{\star} = \argmin_{\boldsymbol{\phi} \in \Phi_{\rho,M}}
    \left\{
    \sum_{s=1}^{t-1}\ell(p_{\boldsymbol{\phi}},Q_s)
    -\ln\pi(\boldsymbol{\phi})
    \right\}.
\end{equation}
The timing notation is prequential: $\boldsymbol{\phi}_t$ is inferred given collected information from the sessions $Q_1,\dots,Q_{t-1}$, before $Q_t$ is observed. With the regulariser equal to the negative log-prior, $\boldsymbol{\phi}_t^{\star}$ is exactly the constrained maximum a posteriori (MAP) estimate based on $\mathcal D_{t-1}$; by the convexity noted in Section~\ref{sec:BAYES}, computing it is a convex optimization problem.

For shape parameters $\alpha,\alpha_{\Delta}>1$ (Section~\ref{sec:hyper}), the regulariser is strictly convex and provides a fixed amount of curvature, ensuring that every FTRL update has a unique solution even when the first observations are weakly informative. In order to achieve logarithmic regret, we make the theoretical assumption that observations keep adding information in all parameter directions. Let $\ell_t(\boldsymbol{\phi})=\ell(p_{\boldsymbol{\phi}},Q_t)$ denote the loss incurred on session $t$.

\begin{assumption}[Persistent curvature]
\label{ass:persistent_curvature}
There exist constants $\mu_{\rho,M}>0$ and $\kappa_{\rho,M}>0$, independent of the time horizon, such that, for every $t\geqslant1$, the regularised cumulative objective
$
F_t(\boldsymbol{\phi})
=
-\ln\pi(\boldsymbol{\phi})
+\displaystyle\sum_{s=1}^{t}\ell_s(\boldsymbol{\phi})
$
is $(\mu_{\rho,M}+\kappa_{\rho,M}n_t)$-strongly convex on $\Phi_{\rho,M}$, wherever it is finite.
\end{assumption}

Here $\mu_{\rho,M}$ is the initial curvature supplied by the regulariser: for $\alpha,\alpha_{\Delta}>1$ one may take $\mu_{\rho,M}=\min\{\alpha-1,\alpha_{\Delta}-1\}/M^{2}$, since on $\Phi_{\rho,M}$ the Hessian of $-\ln\pi$ is $\mathrm{diag}\bigl((\alpha-1)/w_i^{2},\,(\alpha_{\Delta}-1)/(\Delta_m-\rho)^{2}\bigr)$ and every coordinate is at most $M$. The constant $\kappa_{\rho,M}$ measures the minimum additional curvature accumulated per gap observation. 
The prior curvature keeps the estimation problem well posed even when single sessions are uninformative in some directions.

\begin{theorem}[Logarithmic regret for FTRL-DOR]
\label{thm:ftrl_dor_regret}
Assume $\alpha,\alpha_{\Delta}>1$, that the number of gap observations per session is bounded ($C_t\leqslant C_{\max}$, automatic for a finite reference set with $C_{\max}\leqslant|A^R|-1$), and that Assumption~\ref{ass:persistent_curvature} holds. Then there exist finite constants $C^{\mathrm F}_{\rho,M}$ and $A^{\mathrm F}_{\rho,M}$, independent of $T$, such that, for every $T$ with $n_T\geqslant1$,
\begin{equation*}
\sum_{t=1}^{T}\ell_t(\boldsymbol{\phi}_t^{\star})
-
\inf_{\boldsymbol{\phi}\in\Phi_{\rho,M}}
\sum_{t=1}^{T}\ell_t(\boldsymbol{\phi})
\leqslant
C^{\mathrm F}_{\rho,M}
+
A^{\mathrm F}_{\rho,M}\log n_T .
\end{equation*}
\end{theorem}

The proof and explicit expressions for the constants are given in the online supplementary material (Theorem~\ref{thm:appA_ftrl}).

The guarantees of Sections~\ref{sec:BAYES} and~\ref{sec:FTRL} concern the predictive log-loss \eqref{eq:logloss}: they hold for every realised, potentially adversarial, sequence of sessions, they do not require the independence assumption to be true. This property is likely correlated with the robustness such algorithms can produce in prediction and, empirically, in ranking \citep{grillo2025ordinal}. The guarantees produced by such algorithms hold when inference is produced in an \emph{exact} way. The error of finite Monte Carlo sampling is likewise not covered by the exact-inference statements and should be diagnosed separately.

\subsection{The prior hyperparameters}
\label{sec:hyper}
The hyperparameters $(\alpha,\beta,\alpha_{\Delta},\beta_{\Delta})$ of the prior distribution \eqref{eq:prior} can be set in a smart way, and globally, by looking  directly at the regret analysis. The two algorithms operate in different ways, and are favoured by different choices of shape parameter $\alpha$.

For BAYES-DOR, the shape condition of Theorem~\ref{thm:bayes_dor_regret} requires $\alpha,\alpha_{\Delta}\geqslant1$, and the leading regret constant $A_{\rho}=\alpha N+\alpha_{\Delta}e_{\max}$ is increasing in both shape parameters, so the bound is tightest exactly at the boundary $\alpha=\alpha_{\Delta}=1$. This means also that the best guarantee is attained by the least informative allowed prior.

For FTRL-DOR, the curvature of the regulariser $\Omega(\boldsymbol{\phi})=-\ln\pi(\boldsymbol{\phi})$ feeds Assumption~\ref{ass:persistent_curvature}. The regulariser contributes with a Hessian term $(\alpha-1)/w_i^{2}$ in the weight directions and $(\alpha_{\Delta}-1)/(\Delta_m-\rho)^{2}$ in the threshold directions, so the boundary shapes $\alpha=\alpha_{\Delta}=1$, optimal for BAYES-DOR, would make $\Omega$ merely linear, with no curvature of its own. The value $\alpha=\alpha_{\Delta}=2$ is the smallest integer shape that turns $\Omega$ into a strictly convex self-concordant barrier, with a unique minimiser in the interior of the feasible set, bounded away from the boundaries $w_i=0$ and $\Delta_m=\rho$.

The rate parameters ($\beta$) play no fundamental role, and they only affect additive constants of the bounds. Because of that, no value is theoretically preferred. 
We fix the neutral unit rates $\beta=\beta_{\Delta}=1$: for BAYES-DOR, every weight then has prior mean one (and every increment prior mean $1+\rho$); for FTRL-DOR, the mode of the regulariser lies at $w_i=1$ and $\Delta_m=1+\rho$.
These are the values used throughout Sections~\ref{sec:simulation} and~\ref{sec:example}.

\section{Few-shot B-DOR}
\label{sec:algorithm}

The inference mechanism of Section~\ref{sec:b_dor} is naturally sequential: the
posterior obtained after one elicitation step plays the role of the prior for the
next one ~\eqref{eq:posterior_recursive}. This makes B-DOR a practical \emph{few-shots} elicitation procedure. Rather than
asking the DM to rank a single, large set of reference
alternatives in one session, which is a cognitively demanding and error-prone task when more than a few alternatives are involved, the elicitation process is spread over
a small number of short sessions. 

\begin{algorithm}[h]
\LinesNumbered
\SetKwInput{KwInput}{Input}
\SetKwInput{KwOutput}{Output}
\caption{Few-shot Bayesian-DOR}
\label{eq:algo}
\KwInput{(i) A performance matrix for $A^R \subseteq A$; (ii) $\gamma_j$ for each criterion $g_j$; (iii)  $e_{max}$.}
\KwOutput{ (BAYES-DOR): Ranking samples $\Phi_T^R = \{\boldsymbol{\phi}_T^1, \dots , \boldsymbol{\phi}_T^R\}$. (FTRL-DOR): one ranking $\boldsymbol{\phi}_T^{\star}$.
}

Set $t=1$. Take any subset $S_1\subseteq A^R$\;

The DM distributes $a \in S_1$ into levels $\{L_{1,1}, \dots, L_{1,l_t}\}$ and places blank
cards between levels\;

Convert this information into the ordinal observations $Q_1$
(Section~\ref{sec:bdor_data}) and set $\mathcal{D}_1=\{Q_1\}$\;

Compute the posterior $\pi(\boldsymbol{\phi}\mid\mathcal{D}_1)\propto
P(Q_1\mid\boldsymbol{\phi})\,\pi(\boldsymbol{\phi})$\;

\For{$t = 2,\ldots,T$}{
  Take any subset $S_t\subseteq A^R$\;
  The DM distributes $a \in S_t$ into levels $\{L_{t,1}, \dots, L_{t,l_t}\}$ and places blank cards between levels\;
  Convert this information into the ordinal observations $Q_t$ and set $\mathcal{D}_t=\mathcal{D}_{t-1}\cup\{Q_t\}$\;
  Update the posterior $\pi(\boldsymbol{\phi}\mid\mathcal{D}_t)\propto P(Q_t\mid\boldsymbol{\phi})\ \pi(\boldsymbol{\phi}\mid\mathcal{D}_{t-1})$ \;
} \
(BAYES-DOR): Draw $\Phi_T^R = \{\boldsymbol{\phi}_T^1, \dots , \boldsymbol{\phi}_T^R\}$ from $\pi(\boldsymbol{\phi}\mid\mathcal{D}_T)$; \newline
(FTRL-DOR): Estimate $\boldsymbol{\phi}_T^{\star} = \argmax_{\boldsymbol{\phi} \in \Phi_{\rho,M}} \{\pi(\boldsymbol{\phi}\mid\mathcal{D}_T)\}$\;
\Return (BAYES-DOR): A set of rankings $\Phi_T^R$; (FTRL-DOR): One ranking $\boldsymbol{\phi}_T^{\star}$.
\end{algorithm}

The complete procedure is summarised in Algorithm~\ref{eq:algo}, and a step-by-step description of it follows.
 
\begin{description}
\item[Input and Overview.] The input required by the algorithm includes the elements that define the problem: a set of reference alternatives $A^R$ along with their value scores relative to each criterion (performance matrix); the number of internal characteristic points $\gamma_j$ that define the structure of the piecewise marginal value function for each criterion $g_j$ and the allowed maximum number of cards $e_{max}$ between consecutive levels. Larger $e_{max}$ means having a higher resolution function that maps a continuous preference strength to a discrete card count, capable to store more information. The choice of $e_{max}$, though, is bounded above by any computational requirement that depends on the adopted inference method. Section~\ref{sec:simulation} provides such details for Hamiltonian Monte Carlo.
The output is a recommendation for \emph{all}
alternatives in $A$, not only those shown to the DM. There is no rule for choosing the subsets $S_t$ (they can even be chosen by the DM), and may overlap across time sessions.
 
\item[One session (lines 1-3 and 6-8).] Each session is a standard
deck-of-cards exercise on a small subset $S_t$: the DM distributes the alternatives
into levels and places blank cards between consecutive levels. This statement
is converted, as in Section~\ref{sec:bdor_data}, into the gap observations
$Q_t$: one (direction, cards) response per pair of consecutive levels.
 
\item[Learning across sessions (lines 4 and 9).] The posterior computed after
a session becomes the prior of the next one, following the recursion
\eqref{eq:posterior_recursive}. This is the core of the procedure: a new
session does not restart the inference, it refines it. At the first session
this role is played by the prior \eqref{eq:prior}, which keeps the problem
well posed while data are still scarce. Because the model of
Section~\ref{sec:bdor_clm} assigns positive probability to every possible
answer, an occasional inconsistent judgement shifts the posterior a little;
it does not break the procedure.
 
\item[Output (lines 10-11).] Every parameter vector $\boldsymbol{\phi}$
defines, through \eqref{eq:utility}, a value $U(a)$ for every $a\in A$, and
therefore a complete ranking of $A$. BAYES-DOR draws $R$ samples from the posterior through Monte Carlo methods and thus returns $R$ rankings; their spread is the
uncertainty left after $T$ sessions, summarised by the RAI \eqref{eq:RAI} and the PWI \eqref{eq:PWI}. FTRL-DOR
returns the single constrained MAP estimate, and hence one
ranking, at a much lower computational cost. In both cases the posterior
density is available in closed form up to a normalising constant; turning it
into expectations, credible statements or the indices above, however,
requires the samples, and is therefore a BAYES-DOR output. Notice that such output can be arbitrarily produced at each $t$ after the posterior updating step.
\end{description}

\section{Experimental analysis}
\label{sec:simulation}
Having established the sequential inference framework and its theoretical guarantees, we now explore the algorithms in a practical implementation to test their robustness and performance. We build a \emph{ground-truth} model, and we use it to generate data. We then feed the data to the algorithms and later evaluate their ability to reproduce \emph{ground-truth} behaviour. The section addresses five questions: (i) how accurately do BAYES-DOR and FTRL-DOR recover the DM's true preferences from a finite and possibly noisy stream of deck-of-cards observations, and \emph{which} characteristics of the elicitation problem drive that accuracy (Section~\ref{sec:sim_results});
(ii) how much of the performance is attributed to the \emph{cards} information (also Section~\ref{sec:sim_results}); (iii) how do the proposed estimators compare with the DOR method and its SMAA-DOR (Section~\ref{sec:sim_baselines}); (iv) whether the observed sequential predictive behavior matches the regret guarantees of Section~\ref{sec:b_dor} (Section~\ref{sec:sim_regret});
(v) whether the algorithms exploit robustness when the DM distributes alternatives into the same levels, that is, declares ties (Section~\ref{sec:sim_sets}).
To address all five of them we run a large Monte Carlo experiment over combinations of the factors defining a single ranking problem. 

\subsection{Simulation design}
\label{sec:sim_design}

\paragraph{Performance matrices and Value functions} Experiments are replicated several times, and each replication is associated with a different performance matrix (a matrix of scores on criteria for several alternatives) drawn at random: every alternative $a_i$ is described by $m$ criteria, and each performance score $g_j(a_i)$ is sampled independently and uniformly in $[0,1]$. The DM's \emph{true} preferences are represented by an additive value function with exponential marginal values,
\begin{equation*}
  U^{\text{true}}(a_i) \;=\; \sum_{j=1}^{m} u_j^{\text{true}}\!\big(g_j(a_i)\big),
  \qquad
  u_j^{\text{true}}(x) \;=\; w_j\,\frac{1-e^{-c_j x}}{1-e^{-c_j}},
  \quad j=1,\ldots,m,
\end{equation*}
where the weight vector $\bm{w}=(w_1,\ldots,w_m)$ is drawn from a uniform Dirichlet distribution (so that $w_j\geqslant 0$ and $\sum_j w_j = 1$) and each curvature parameter $c_j$ is drawn uniformly in $[-10,10]$, independently across criteria and replications. The sign of $c_j$ controls the concavity/convexity of the marginal value, so the set of true models spans a broad range of value functions. The true ranking of the alternatives is obtained by sorting them according to $U^{\text{true}}$. On the inference side, the value function is \emph{not} assumed to be exponential: it is estimated through piecewise-linear functions with $\gamma_j = 3$, meaning four equally spaced characteristic points per criterion introduced in Section~\ref{sec:basic_concept}, so that the model is always misspecified with respect to the data-generating process, as it would possibly be in any real application.

\paragraph{Data generation process} Preference information is generated sequentially over $T$ steps. At each step $t$ a synthetic DM is shown a subset $S_t$ containing $k_t$ reference alternatives drawn at random and answers with the ordinal observation $Q_t$ formalised in Section~\ref{sec:bdor_data}. In all experiments, except the one in the dedicated study of Section \ref{sec:sim_sets}, the DM never declares ties: every level is a singleton, and collected data per session is a ranking of the subset, along with cards between alternatives. Because the criterion weights sum to one, each value $U^{\text{true}}(a_i)$ lies in $[0,1]$. 
The alternatives $a$ in each $S_t$ are sorted by their true evaluation $U(a)$, and the number of blank cards inserted between two consecutive alternatives $a\succ b$ is generated by multiplying their evaluation difference by $e_{\max}=5$ and rounding to the nearest integer:
\begin{equation*}
  \operatorname{cards}(a,b)
  \;=\;
  \left\lfloor\, e_{\max}\cdot\left(U^{\text{true}}(a)-U^{\text{true}}(b)\right)\,\right\rceil
  \;\in\;\{0,1,\ldots,e_{\max}\},
\end{equation*}
where $\lfloor\cdot\rceil$ denotes rounding to the nearest integer.  
The larger the gap between two consecutive alternatives, the more blank cards separate them. This simulates a DM that maps preference strength to card counts linearly. The specific choice $e_{max} = 5$ keeps the number of parameters to estimate low, and make inference fast. Higher values can only improve performance, as shown in Section \ref{sec:sim_emax}. 

\paragraph{Injecting inconsistency} A perfectly rational DM would always return the ranking and the blank-card gaps induced by $U^{\text{true}}$. Real DMs, however, are imprecise. We reproduce this through a  utility value noise model. Before the alternatives of a step are ranked, the $e_{\max}$-scaled true evaluation of each displayed alternative ($e_{\max} \cdot U^{true}(a_i)$) is perturbed by an independent Gaussian shock \citep{thurstone1927law,maydeu2005structural},
\begin{equation}
\label{eq:noisy_utility}
  \widetilde U(a_i)\;=\;e_{\max}\cdot U^{\text{true}}(a_i)+\varepsilon_i,
  \qquad
  \varepsilon_i \stackrel{\text{i.i.d.}}{\sim}\mathcal N\!\big(0,\,\sigma^{2}/2\big),
\end{equation}
where the per-alternative variance $\sigma^{2}/2$ is chosen so that the difference $\widetilde U(a)-\widetilde U(b)$ between any two displayed alternatives has variance $\sigma^{2}$. The DM then ranks the subset according to $\widetilde U$ and, for two consecutive alternatives $a\succ b$ in this noisy ranking, declares the card count
\begin{equation*}
  \operatorname{cards}(a,b)\;=\;\min\!\Big\{\,e_{\max},\;\;\big\lfloor\,\widetilde U(a)-\widetilde U(b)\,\big\rceil\Big\},
\end{equation*}

Introducing inconsistencies with this method, means that mistakes are expected to be rare when the comparisons are easy (few alternatives, clearly separated evaluations) and become more frequent as the task includes more, less-separated alternatives.

The amount of noise is governed by the single factor $F_{\text{inc}}$, which fixes the noise scale $\sigma$ in \eqref{eq:noisy_utility}. Consider a displayed pair with true scaled gap $\mu=e_{\max}\big(U^{\text{true}}(a)-U^{\text{true}}(b)\big)$; under the noise model its elicited gap is $v\sim\mathcal N(\mu,\sigma^{2})$, so the elicited card count $\lfloor v\rceil$ coincides with the true one $\lfloor\mu\rceil$ with probability
\begin{equation*}
  q(\mu,\sigma)\;=\;\varphi\!\Big(\tfrac{\lfloor\mu\rceil+\frac12-\mu}{\sigma}\Big)-\varphi\!\Big(\tfrac{\lfloor\mu\rceil-\frac12-\mu}{\sigma}\Big),
\end{equation*}
where $\varphi$ is the standard normal cumulative distribution function. For each problem configuration we set $\sigma$ as the solution of
\begin{equation}
\label{eq:finc_calibration}
  F_{\text{inc}}\;=\;1-\mathbb E_{\mu}\big[\,q(\mu,\sigma)\,\big],
\end{equation}
where the expectation is taken over the empirical distribution of true scaled gaps $\mu$ between randomly paired alternatives, estimated once per problem size on a large Monte Carlo calibration sample. In other words, $\sigma$ is tuned so that, on average, a fraction $F_{\text{inc}}$ of the adjacent gap observations carries a number of blank cards different from the one implied by the true evaluations.

Because a perturbed gap may report a wrong number of blank cards without actually reversing the underlying order, the nominal level $F_{\text{inc}}$ is a conservative upper bound on the proportion of genuinely \emph{flipped} preferences. To make the amount of injected noise transparent, we measured directly the ratio of inverted pairwise comparisons on the generated elicitation data. At the nominal levels $F_{inc}=\{0,15,35,50\}\%$, the realized fraction of reversed comparisons is about $\{0.0,\,7.7,\,15.4,\,21.5\}\%$ when all $\binom{k}{2}$ comparisons declared by each ranking are counted, and $\{0.0,\,13.6,\,23.6,\,29.7\}\%$ when restricted to the adjacent comparisons. The portion of flipped preferences, averaged over all generated data is reported in Table~\ref{tab:realized_flips}. We additionally observe that it grows mildly with the number of criteria (from $21.0\%$ at $m=3$ to $23.9\%$ at $m=9$), while it grows more visibly with the subset size, from $18.7\%$ at $k=3$ to $24.5\%$ at $k=5$. This is because the more alternatives there are to rank in one session, the closer they are in true utility value; therefore, flipping happens more often. We return to this effect in Section~\ref{sec:sim_results}. 

\begin{table}[h!]
\centering
\small
\caption{Realized share of \emph{flipped} pairwise comparisons (preferences deduced from ranking, contradicting the true order) for each nominal inconsistency level $F_{\text{inc}}$.}
\label{tab:realized_flips}
\begin{tabularx}{0.85\textwidth}{l|CCCC}
\toprule
Nominal $F_{\text{inc}}$ & $0\%$ & $15\%$ & $35\%$ & $50\%$\\
\midrule
Flipped comparisons (all $\binom{k}{2}$ deduced pairs) & $0.0\%$ & $7.7\%$ & $15.4\%$ & $21.5\%$\\
Flipped comparisons (adjacent pairs only) & $0.0\%$ & $13.6\%$ & $23.6\%$ & $29.7\%$\\
\bottomrule
\end{tabularx}
\end{table}
 
\paragraph{Inference details} For every experimental configuration we run both estimators of Section~\ref{sec:algorithm}: BAYES-DOR (Section~\ref{sec:BAYES}), whose posterior is explored by Hamiltonian Monte Carlo in its No-U-Turn Sampler (NUTS) variant, and FTRL-DOR (Section~\ref{sec:FTRL}), which computes the regularised maximum-a-posteriori parameter by constrained convex optimisation. Both rely on the prior \eqref{eq:prior}, with $\alpha=\alpha_{\Delta}=\beta=\beta_{\Delta}=1$ for BAYES-DOR and $\alpha=\alpha_{\Delta}=2$, $\beta=\beta_{\Delta}=1$ for FTRL-DOR: these values are not tuned but follow from the regret theory (Section~\ref{sec:hyper}). The threshold floor is $\rho=10^{-6}$, the theory requiring only $\rho>0$: small, yet large enough to keep consecutive thresholds strictly separated in floating-point arithmetic. Each Bayesian update runs the NUTS implementation of NumPyro \citep{phan2019composable} with $200$ warm-up iterations followed by $R=200$ collected draws, from which the rank acceptability and pairwise winning indices are computed; the FTRL convex optimization is solved by L-BFGS-B \citep{byrd1995limited} as provided by SciPy \citep{virtanen2020scipy}, with exact gradients obtained by automatic differentiation in JAX \citep{bradbury2018jax}\footnote{The computational load is light: on a single desktop CPU core, one posterior refresh takes about $1.0$--$1.6$ seconds and a complete FTRL run over a ten-session sequence about $0.4$ seconds, so the full study completes in under 7 hours on a standard workstation.}. Sampler health was monitored on every Bayesian fit: the split-$\widehat{R}$ \citep{vehtari2021rank} of the slowest-mixing parameter exceeded $1.05$ in $0.6\%$ of the runs; meaning such settings are well suited for the $e_{max} = 5$ choice. All runs use deterministic seeds indexed by (configuration, method, replication, session).

\paragraph{Compared methods} Besides the two proposed algorithms, every experimental configuration is run by four other methods, fed with \emph{exactly the same} elicitation data. (i)~The \emph{DOR} is used as a point estimator. (ii)~The SMAA-DOR draws value functions uniformly from the space of models compatible with the declared rankings and card counts. We draw $200$ samples by Hit-and-Run \citep{tervonen2013hit,van2014notes}, matching the posterior sample size of BAYES-DOR (Section~\ref{sec:sim_baselines}). Both methods are implemented exactly as formulated in Section~\ref{sec:dor}. (iii-iv)~A \emph{direction-only} version of our own algorithms, in which the likelihood retains the declared directions but ignores the card counts. This is exactly the ordinal-regression setting of \citet{grillo2025ordinal}. Both the Bayesian and the FTRL variants (BAYES$_{\text{dir}}$, FTRL$_{\text{dir}}$) are tested (Section~\ref{sec:sim_results}).

\paragraph{Experimental grid} We consider a full factorial design over the factors and the levels reported in Table~\ref{tab:grid}; crossing all levels yields $4\times4\times4\times3\times4=768$ cells. The four horizons $T$ are not simulated independently: a replication runs the whole sequence of ten sessions and the inferred recommendation is evaluated after sessions $1$, $3$, $5$ and $10$, so the shorter horizons are prefixes of the longer ones.

\begin{table}[h!tp]
\centering
\small
\begin{tabular}{lcc}
\toprule
Factor & Symbol & Levels\\
\midrule
Number of alternatives & $|A|$ & $10,\ 20,\ 35,\ 50$\\
Number of criteria & $m$ & $3,\ 5,\ 7,\ 9$\\
Number of elicitation steps & $T$ & $1,\ 3,\ 5,\ 10$\\
Number of alternatives ranked at each step (subset size) & $k$ & $3,\ 4,\ 5$\\
Share of inconsistent data & $F_{\text{inc}}$ & $0\%,\ 15\%,\ 35\%,\ 50\%$\\
\bottomrule
\end{tabular}
\caption{Factors of the full factorial simulation design and their levels.}
\label{tab:grid}
\end{table}
Note that we keep the same subset size $k$ at every elicitation step, but this is not an algorithm constraint and can in principle differ from session to session. In our experiments, the data-generating process is identical at every step, but in a real human process the first step is usually the most informative and the least prone to error, so in a real case scenario, there would be reason to give it a different size from the subsequent ones. Every cell of the grid is replicated on $20$ independently generated performance matrices and value functions. All reported performance scores are averages over these replications.

\subsection{Performance metrics}
\label{sec:metrics}
Following \citet{ru2022bayesian}, we evaluate the algorithms using these three metrics:
\paragraph{Average Support for the true Rank (ASR)} By denoting $\text{rank}^{\text{true}}(a_i)$ the rank of alternative $a_i$ according
to the true preference model $U^{\text{true}}$,
\begin{equation}
\label{eq:ASR}
  \text{ASR} \;=\; \frac{1}{n}
  \sum_{i=1}^{n} \text{RAI}\!\left(a_i,\,\text{rank}^{\text{true}}(a_i)\right).
\end{equation}
$\text{ASR}$, therefore, measures how well the inferred model concentrates probability mass on the true ranking of each alternative.
\paragraph{Average Support for true Pairwise comparisons (ASP)}
Let $\mathbb{I}(U^{\text{true}}(a_i)\geqslant U^{\text{true}}(a_j))$ equal $1$ if
alternative $a_i$ is at least as good as $a_j$, and $0$ otherwise.
\begin{equation}
\label{eq:ASP}
  \text{ASP} \;=\; \frac{2}{n(n-1)} \sum_{i=1}^{n}\sum_{\substack{j=1\\j\neq i}}^{n}
  \text{PWI}(a_i,a_j)\cdot \mathbb{I}\left(U^{\text{true}}(a_i)\geqslant U^{\text{true}}(a_j)\right).
\end{equation}
$\text{ASP}$ measures the overall alignment of the inferred pairwise comparisons with those implied by the true preference model.
\paragraph{Acceptability Index of the true Optimal alternative (AIO)}
Let us denote $a^*$ the top-ranked alternative according to $U^{\text{true}}$.
\begin{equation}
  \label{eq:AIO}
  \text{AIO} \;=\; \text{RAI}(a^*, 1),
\end{equation}
$\text{AIO}$ measures the proportion of inferred models for which the top-ranked alternative coincides with the true optimal $a^*$. It focuses on the most decision-relevant quantity: correctly identifying the best
alternative.

All three metrics belong to $[0, 1]$, with higher values indicating better performance of the inferred model. They are computed from the final sample of value functions returned by each method: for BAYES-DOR and SMAA-DOR this is the posterior (respectively, uniform) sample, while for the point estimators (FTRL-DOR and the original DOR) the sample contains a single value function, so the RAI and PWI reduce to $0/1$ indicators and the three metrics score the single inferred ranking. Nonetheless presented scores relative to point estimators are averaged over $20$ repetitions, so fractional scores are still possible.

\subsection{B-DOR performance and robustness}
\label{sec:sim_results}
\begin{figure}[h!]
    \centering
    \includegraphics[width=0.8\textwidth]{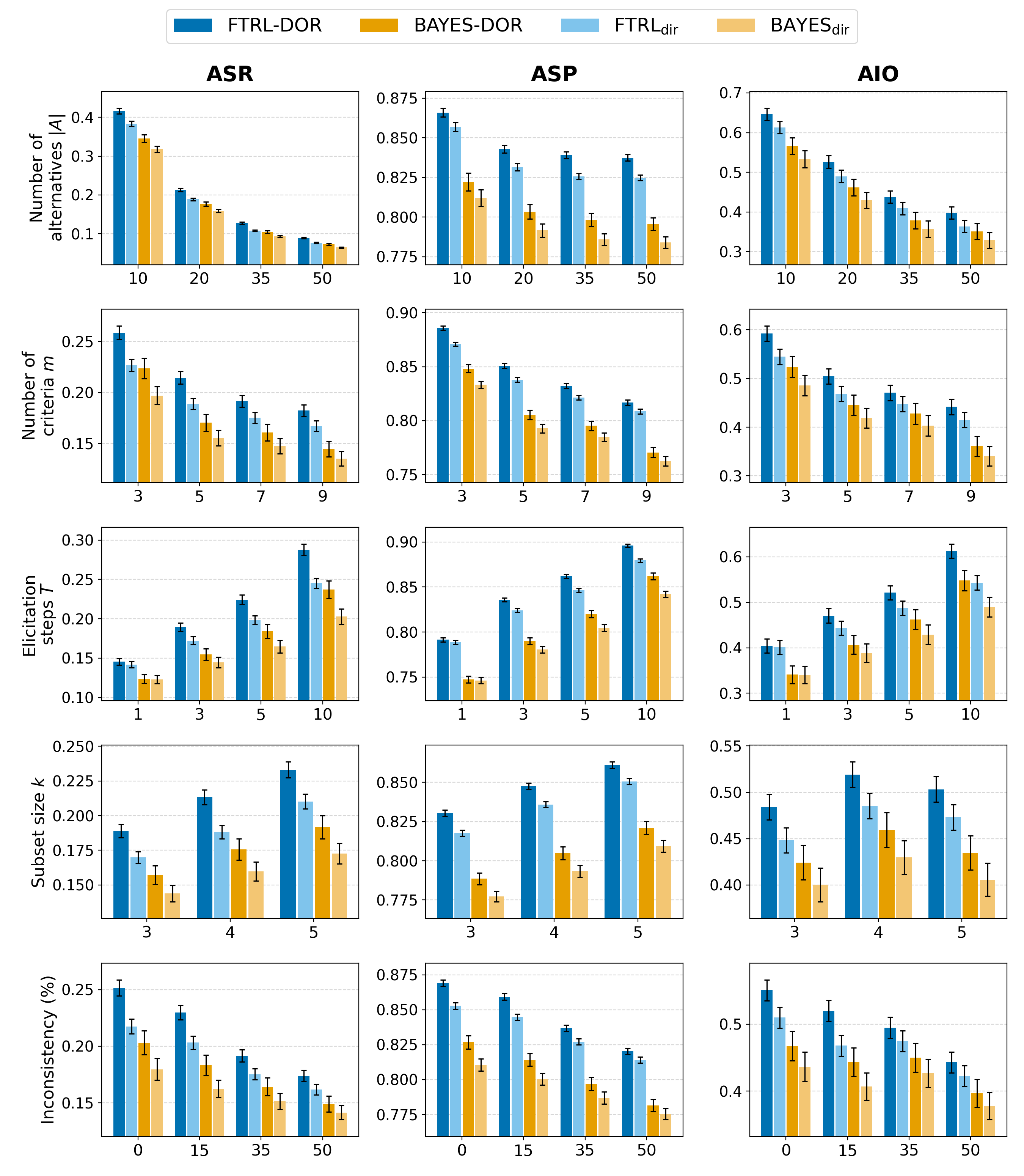}
    \caption{Marginal effect of each experimental factor on the three metrics for the two B-DOR algorithms and their direction-only version.
    Within each plot the bars are sorted from best to worst; error bars are $95\%$ confidence intervals.
    }
    \label{fig:simulation_factors}
\end{figure}

This section evaluates the two B-DOR algorithms (FTRL-DOR and BAYES-DOR), and also compares them with their direction-only versions FTRL$_{\text{dir}}$ and BAYES$_{\text{dir}}$. Experiments are done for all configurations in Table~\ref{tab:grid}. The results are shown in two ways: the upper block of Table~\ref{tab:overall_summary} gives the overall picture: the mean of each metric over the whole grid of configurations, accompanied by the $p$-value of a one-sided paired Wilcoxon signed-rank test, checkinf if the first method scores higher than the second, paired on the same simulation; 
Figure~\ref{fig:simulation_factors} shows the effect of each experimental factor separately: the factor is fixed at some value and the metric is averaged over all the other factors. The mean of each metric when holding one factor fixed to a specific value are collected in Tables~\ref{tab:app_alt}-\ref{tab:app_inc} in the online supplementary material.

\paragraph{Main findings} \begin{itemize}
    \item B-DOR algorithms perform better than their direction-only version, so the introduced probabilistic framework is able to capture the cards information that matter. 
    \item FTRL-DOR is the best-performing algorithm on all three metrics, and the performance order of the four algorithms is identical on every metric: FTRL-DOR, FTRL$_{\text{dir}}$, BAYES-DOR, BAYES$_{\text{dir}}$. 
    \item Whether one should prefer iterating over more sessions $T$ or having larger $k_t$ per session depends entirely on the DM: a more inconsistent DM should prefer larger $T$. 
    \item A finer card scale (larger $e_{\max}$) makes the model capable to store more information and pays off with a consistent and accurate DM, but if the DM's imprecision grows at least linearly with $e_{\max}$, allowing more cards does not improve performance.
\end{itemize}

\begin{table}[h!]
    \centering
    \small
\begin{tabular}{l|ccc}
\toprule
&ASR&ASP&AIO\\
\midrule
FTRL-DOR & $0.212{\scriptstyle\pm0.003}$ & $0.846{\scriptstyle\pm0.001}$ & $0.502{\scriptstyle\pm0.008}$ \\
FTRL$_{\text{dir}}$ & $0.189{\scriptstyle\pm0.003}$ & $0.835{\scriptstyle\pm0.001}$ & $0.469{\scriptstyle\pm0.008}$ \\
BAYES-DOR & $0.175{\scriptstyle\pm0.004}$ & $0.805{\scriptstyle\pm0.002}$ & $0.439{\scriptstyle\pm0.011}$ \\
BAYES$_{\text{dir}}$ & $0.159{\scriptstyle\pm0.004}$ & $0.793{\scriptstyle\pm0.002}$ & $0.412{\scriptstyle\pm0.010}$ \\
\quad FTRL-DOR $>$ FTRL$_{\text{dir}}$ (Wilcoxon $p$) & $<0.001$ & $<0.001$ & $<0.001$ \\
\quad BAYES-DOR $>$ BAYES$_{\text{dir}}$ (Wilcoxon $p$) & $<0.001$ & $<0.001$ & $<0.001$ \\
\midrule
\multicolumn{4}{l}{\itshape First session only ($T=1$), proposed vs existing methods (Section~\ref{sec:sim_baselines})}\\[2pt]
FTRL-DOR & $0.145{\scriptstyle\pm0.004}$ & $0.792{\scriptstyle\pm0.002}$ & $0.404{\scriptstyle\pm0.016}$ \\
DOR & $0.121{\scriptstyle\pm0.004}$ & $0.750{\scriptstyle\pm0.003}$ & $0.253{\scriptstyle\pm0.014}$ \\
BAYES-DOR & $0.123{\scriptstyle\pm0.003}$ & $0.748{\scriptstyle\pm0.002}$ & $0.330{\scriptstyle\pm0.010}$ \\
SMAA-DOR & $0.117{\scriptstyle\pm0.004}$ & $0.718{\scriptstyle\pm0.004}$ & $0.267{\scriptstyle\pm0.013}$ \\
\quad FTRL-DOR $>$ DOR (Wilcoxon $p$) & $<0.001$ & $<0.001$ & $<0.001$ \\
\quad BAYES-DOR $>$ SMAA-DOR (Wilcoxon $p$) & $<0.001$ & $<0.001$ & $<0.001$ \\
\bottomrule
\end{tabular}
    \caption{Upper block: mean of each metric over the whole simulation grid, with $95\%$ confidence intervals. Lower block: the same means after the first elicitation session only ($T=1$), for the proposed estimators and the DOR and SMAA-DOR. Both blocks report the $p$-value of a one-sided paired Wilcoxon signed-rank test of the hypothesis that the first method scores higher than the second.}
    \label{tab:overall_summary}
\end{table}

\paragraph{Overall comparison} Averaged over the whole grid (Table~\ref{tab:overall_summary}, upper block), the four algorithms' performance are aligned on every metric: FTRL-DOR gets the highest scores ($\text{ASR}=0.212$, $\text{ASP}=0.846$, $\text{AIO}=0.502$), then FTRL$_{\text{dir}}$ ($0.189$, $0.835$, $0.469$), then BAYES-DOR ($0.175$, $0.805$, $0.439$), then BAYES$_{\text{dir}}$ ($0.159$, $0.793$, $0.412$). Two main results follow. First, the FTRL family is ahead of the Bayesian family. In fact, FTRL-DOR returns one best-fitting model, whereas BAYES-DOR more robustly draws a finite sample of $R=200$ rankings from the posterior, but pays a cost in performance. Second, as previously stated, B-DOR algorithms perform better than their direction-only version.

\paragraph{Number of alternatives ($|A|$)} All algorithms lose accuracy as $|A|$ grows (first row of Figure~\ref{fig:simulation_factors}; Table~\ref{tab:app_alt} in the online supplementary material). The loss is large on ASR,
intermediate on AIO, 
and small on ASP. With the same amount of preference information, guessing the exact rank of every alternative in a larger set is much harder, and it becomes less difficult when you have to guess only the top-ranked alternative, while getting the direction of the pairwise comparisons right degrades only mildly. The performance ordering of the four algorithms is preserved at every $|A|$; The Wilcoxon tests of Table~\ref{tab:app_alt} confirm that the ordering is not only preserved but significant everywhere: even at |A|=50, where the ASR bars are visually indistinguishable, each algorithm is still ahead of the next one with p<0.01 on every metric.

\paragraph{Number of criteria ($m$)} Similar to alternatives, adding criteria to the MCDA problem degrades performance for all algorithms, although moderately (second row of Figure \ref{fig:simulation_factors}; Table~\ref{tab:app_crit} in the online supplementary material): for FTRL-DOR, ASR falls from $0.259$ at $m=3$ to $0.182$ at $m=9$, and the other three algorithms drop by similar fractions. All algorithms preserve the robustness they are designed for: there is no value at which any of them breaks down. The performance ordering of the four algorithms is again unchanged at every $m$, with the only exception being, for ASR: at $m=3$, FTRL$_{\text{dir}}$ and BAYES-DOR are nearly tied. The Wilcoxon tests of Table~\ref{tab:app_crit} confirm the gain of B-DOR algorithms over the direction-only version: at every $m$, for both estimators ($p<0.001$).

\paragraph{Number of elicitation steps ($T$)} More sessions improve performance of every algorithm on every metric (third row of Figure \ref{fig:simulation_factors}; Table~\ref{tab:app_steps} in the online supplementary material), and most of the benefit arrives early. For FTRL-DOR, ASR grows from $0.145$ after a single session to $0.190$ after three and to $0.288$ after ten. Two extra sessions are thus worth about a third of everything that ten sessions deliver. At $T=3$ one gets $31\%$ of the ASR gain at $T=10$, $42\%$ for ASP and $32\%$ for AIO. The practical message is that a long interaction is not needed: a few rounds already give a good picture of the DM's preferences, and further rounds keep refining it at a slower pace. The same pattern holds for all four algorithms.

After only a single session B-DOR and its direction-only version are almost identical (for the FTRL pair, ASR $0.145$ against $0.142$), 
and the paired tests are inconclusive on two metrics: $p=0.012$ on ASR and $p=0.181$ on AIO for FTRL (similar for BAYES). From the third session onward the picture is clearer: both FTRL-DOR and BAYES-DOR beat their direction-only versions on every metric with $p<0.001$, at every subsequent horizon, and the advantage keeps growing.

\paragraph{Subset size ($k$)} Ranking more alternatives per session generally improves performance (fourth row of Figure \ref{fig:simulation_factors}; Table~\ref{tab:app_sub} in the online supplementary material): from $k=3$ to $k=5$, ASR moves from $0.189$ to $0.233$ for FTRL-DOR and from $0.157$ to $0.192$ for BAYES-DOR, and ASR and ASP generally improve for every algorithm. 
The only non-monotone pattern concerns AIO, which peaks at $k=4$ (for FTRL-DOR, $0.484/0.519/0.503$ at $k=3/4/5$). We recall that, in this study, scores are averaged over all factors of Table~\ref{tab:grid}, except $k$. Averaging over $F_{inc}$ is what causes the drop: when re-computing scores only with $F_{inc} = 0\%$ data, AIO grows with $k$ like the other two metrics ($0.425/0.465/0.513$ for BAYES-DOR). The cause is that the alternatives in a session are closer in true evaluation when the subset is larger (their mean distance falls from $0.120$ at $k=3$ to $0.090$ at $k=5$) and closer alternatives are reversed more easily. AIO is the metric that suffers the most because it is the only one that is not computed through an average: it scores a single event (whether the true best alternative is ranked first), and one preference flip involving that alternative destroys it completely. In practice, larger subsets are worthwhile with a reliable DM, while with a very inconsistent one it is better to keep them small.

The additional cards information pay off at every subset size: both B-DOR algorithms are ahead of their direction-only versions with $p<0.001$ at $k=3$, $4$ and $5$, on all three metrics.

\paragraph{Larger sets or more sessions?}
For B-DOR algorithms, the atomic element of information that matters is the \emph{gap observation}. The total amount of them collected at session $T$ is represented by equation~\eqref{eq:nT}, which in our experiments simplifies to $n_T=T(k-1)$. This is linear both in $k$ and $T$, so having more sessions or larger ones is statistically equivalent. Two differences still matter: first, when running over sessions, alternatives can repeat, making it easier for a less decided DM, although possibly providing a lower amount of information for the final ranking task; second, whether to prefer more sessions rather than larger ones depends entirely on the DM and its ability to provide clean information even at large $k_t$. In our experiments we observe that when noise is introduced using the Thurstonian noise model described in Section~\ref{sec:sim_design}, it is preferred to have more sessions than larger ones even at our nominal $F_{inc} = 15\%$ level.

\paragraph{Inconsistent preferences ($F_{\text{inc}}$)} Noisy answers degrade the scores almost linearly with $F_{inc}$ for all algorithms and across all metrics (fifth row of Figure \ref{fig:simulation_factors}; Table~\ref{tab:app_inc} in the online supplementary material). From a fully consistent DM to $F_{\text{inc}}=50\%$, the ASR of FTRL-DOR falls from $0.252$ to $0.174$ and that of BAYES-DOR from $0.203$ to $0.149$. The drop with inconsistency is small: for example FTRL-DOR reproduces $82\%$ of the true pairwise comparisons, against $87\%$ with a perfectly consistent DM. The inferred model is therefore a robust representation of the true one. B-DOR algorithms keep their advantage against the direction-only versions at every noise level ($p<0.001$ on all three metrics), even if the advantage shrinks as the DM gets noisier.

\subsection{A discussion on $e_{\max}$}\label{sec:sim_emax} All the experiments above fix $e_{\max}=5$. Increasing it, would also increase the dimensionality of $\boldsymbol{\Delta}$ (see ~\eqref{eq:thresholds}) yielding a more complex model that is capable to store more information about the DM response to preference strength. Whether this also makes the recommendations better depends on the DM and its ability to be consistent. We repeated one experiment on configuration ($|A|=10$, $m=3$, $k=3$, $T=10$) with $e_{\max}=50$, over the $20$ replications, and setting the number of samples drawn by the BAYES-DOR to $R=2000$. With a fully consistent DM the extra resolution pays off: at $T=10$ the ASR of BAYES-DOR rises from $0.602$ to $0.734$ and its ASP from $0.931$ to $0.960$ (paired Wilcoxon $p=0.002$ on both), with gains also for FTRL-DOR (ASR $0.740$ against $0.820$). This scenario, however, assumes that the DM places single cards on a fifty-card scale with absolute precision. If instead the DM's imprecision scales with the maximum number of allowed cards, the benefit disappears: repeating the experiment with inconsistent preferences, and multiplying the noise scale $\sigma$ of \eqref{eq:noisy_utility} by ten (the ratio between the new $e_{\max}$ and the old one), ASR improves by only $+0.060$ for FTRL-DOR and $+0.037$ for BAYES-DOR, neither significant ($p\geqslant0.19$). A larger $e_{\max}$ therefore improves performance only if the error associated with the \emph{blank-cards} random variable grows sub-linearly with $e_{\max}$.

\subsection{Comparison with the existing DOR methods}\label{sec:sim_baselines} 
\begin{figure}[h!]    
\centering    
\includegraphics[width=0.8\textwidth]{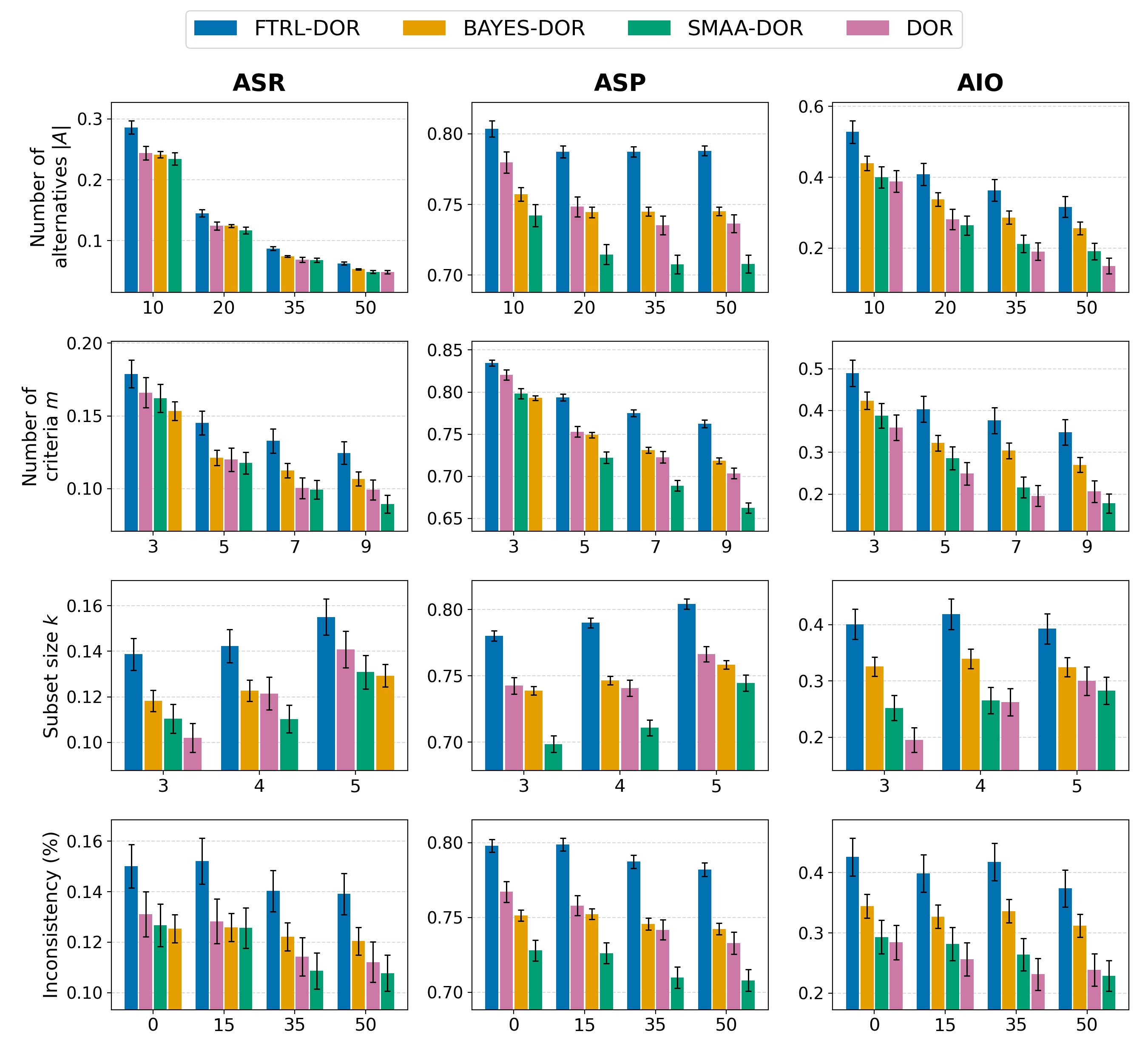}
\caption{Marginal effect of each experimental factor on the three metrics, for the two B-DOR algorithms, SMAA-DOR and the original DOR. Within each level group the bars are sorted from best to worst; error bars are $95\%$ confidence intervals. Single-session comparison ($T=1$).
}    
\label{fig:baselines_t1}
\end{figure}

We now compare the proposed estimators with the two existing methods of the DOR family, SMAA-DOR and the original DOR. Both are one-shot methods: they turn one given block of preference information (one session) into a recommendation, and they specify neither how to revise it when further sessions arrive nor what to do when observations collected in different sessions contradict each other and empty the compatible space of solutions. To keep the comparison fair, we run it in the setting for which the two methods were designed for: all four methods are evaluated after the first elicitation session ($T=1$), on identical data, across the whole grid of the remaining factors. The lower block of Table~\ref{tab:overall_summary} reports the overall means and the significance tests; Figure~\ref{fig:baselines_t1} the marginal effect of each factor; Tables~\ref{tab:app_t1_alt}-\ref{tab:app_t1_inc} in the online supplementary material the fixed-factor detailed results. 

\paragraph{Overall picture} FTRL-DOR is the most accurate method on all three metrics after a single session ($\text{ASR}=0.145$, $\text{ASP}=0.792$, $\text{AIO}=0.404$; Table~\ref{tab:overall_summary}, lower block). For the remaining algorithms the ordering depends on the metric. On AIO the order (after FTRL-DOR) is BAYES-DOR ($0.330$), then SMAA-DOR ($0.267$), then DOR ($0.253$). On ASR, BAYES-DOR ($0.123$) is followed by the two existing methods almost tied ($0.121$ for DOR, $0.117$ for SMAA-DOR). When looking at ASP, DOR climbs to a near-tie with BAYES-DOR for second place ($0.750$ against $0.748$) while SMAA-DOR stays fourth ($0.718$). The Wilcoxon tests of Table~\ref{tab:overall_summary} (one-sided and paired on the same simulated replications) check whether the new algorithms perform better than existing methods of the same kind ( FTRL-DOR against DOR, and BAYES-DOR against SMAA-DOR): both win with significant advantage $p<0.001$ on every metric. 

DOR is designed to maximally separate the alternatives, and this happens at a
corner (vertex) of the polytope (the space of feasible solutions): it
concentrates the weights on the few criteria that most distinguish the
alternatives shown in the session (about $80\%$ of the components get zero
weight on average), so its direction fits the declared comparisons sharply and its ASP
performance is good, behind FTRL-DOR only. SMAA-DOR returns vectors drawn
uniformly from the polytope's interior; its weights
are spread over all the criteria, so it identifies the true best alternative
more often. This is why the two existing methods swap order between ASP and
AIO, while on ASR, which is sensitive to both effects, they end up almost
tied. FTRL-DOR and BAYES-DOR avoid the trade-off: the regularized-likelihood
inference mechanism outputs parameter vectors that \emph{more likely}
reproduce preferences, but also stay spread across the criteria.

\paragraph{Problem size ($|A|$ and $m$)} When looking at performance isolating the influence of factors $|A|$ and $m$, the general conclusions of the overall picture basically hold (first two rows of Figure~\ref{fig:baselines_t1}; Tables~\ref{tab:app_t1_alt} and~\ref{tab:app_t1_crit} in the online supplementary material), with few considerations worth of notice. For example, on ASP the two point estimators are closest at $|A|=10$ ($0.804$ against $0.780$), and the gap widens as $|A|$ grows: DOR slides to $0.736$ at $|A|=50$ while FTRL-DOR stays extremely robust ($0.788$). The same behaviour can be observed for the sampling variants. FTRL-DOR and BAYES-DOR exploit not only better performance with respect to DOR and SMAA-DOR respectively, but also greater robustness to the increasing complexity of the ranking problem. The number of criteria fixes the dimension of the polytope, and it is the factor where classic DOR methods suffer most. At $m=3$ the polytope is low-dimensional, every compatible model is close to every other, and both DOR methods are at their best.
Here both existing methods overtake BAYES-DOR on ASR ($0.166$ for DOR and $0.162$ for SMAA-DOR against $0.153$), and SMAA-DOR does it on ASP as well ($0.798$ against $0.793$; the paired tests are inconclusive, $p_{\mathrm{B>S}}=0.077$ on ASR and $0.890$ on ASP in Table~\ref{tab:app_t1_crit}), while FTRL-DOR keeps the lead on every metric. From $m=3$ to $m=9$ SMAA-DOR's ASR and AIO roughly halve ($0.162$ to $0.089$; $0.388$ to $0.178$) and its ASP loses more than any other method ($0.798$ to $0.663$), while FTRL-DOR loses about $30\%$ of its ASR. From $m=5$ onwards BAYES-DOR outperforms SMAA-DOR on every metric with $p<0.001$: the advantage of the likelihood-based sampler over the uniform one grows when the polytope grows.

\paragraph{Subset size ($k$) and inconsistency ($F_{\text{inc}}$)} At $T=1$ the subset size is the only lever on the amount of collected information, and indeed ASR and ASP improve with $k$ for all four methods (Table~\ref{tab:app_t1_sub} in the online supplementary material). AIO partially contradicts the information principle: FTRL-DOR and BAYES-DOR peak at $k=4$ but have a lower score at $k=5$, while the two existing methods keep improving with $k$ (DOR from $0.195$ to $0.300$). The performance drop of the B-DOR algorithms happens because larger subsets contain alternatives that are closer in true utility value, hence more often flipped by the DM (Section~\ref{sec:sim_results}). The information required to spot the true best alternative does not grow with $k$ fast enough to balance this mechanism, so AIO gets a ceiling reachable value. DOR and SMAA-DOR start from lower AIO scores, so the extra information given by a larger subset still causes an increase. Inconsistency only slightly perturbs any method after one session (FTRL-DOR's ASR moves from $0.150$ at $F_{\text{inc}}=0\%$ to $0.139$ at $50\%$, and the two existing methods lose slightly more, DOR from $0.131$ to $0.112$; Table~\ref{tab:app_t1_inc} in the online supplementary material). Inconsistency injection is the most forgiving setting for both DOR and SMAA-DOR, since at $T=1$ the collected information does not allow for cyclic contradictions across sessions. Even in this case the likelihood-based estimators are ahead.

\subsection{Experiments with tied alternatives}
\label{sec:sim_sets}
\begin{figure}[h!]
    \centering
    \includegraphics[width=\textwidth]{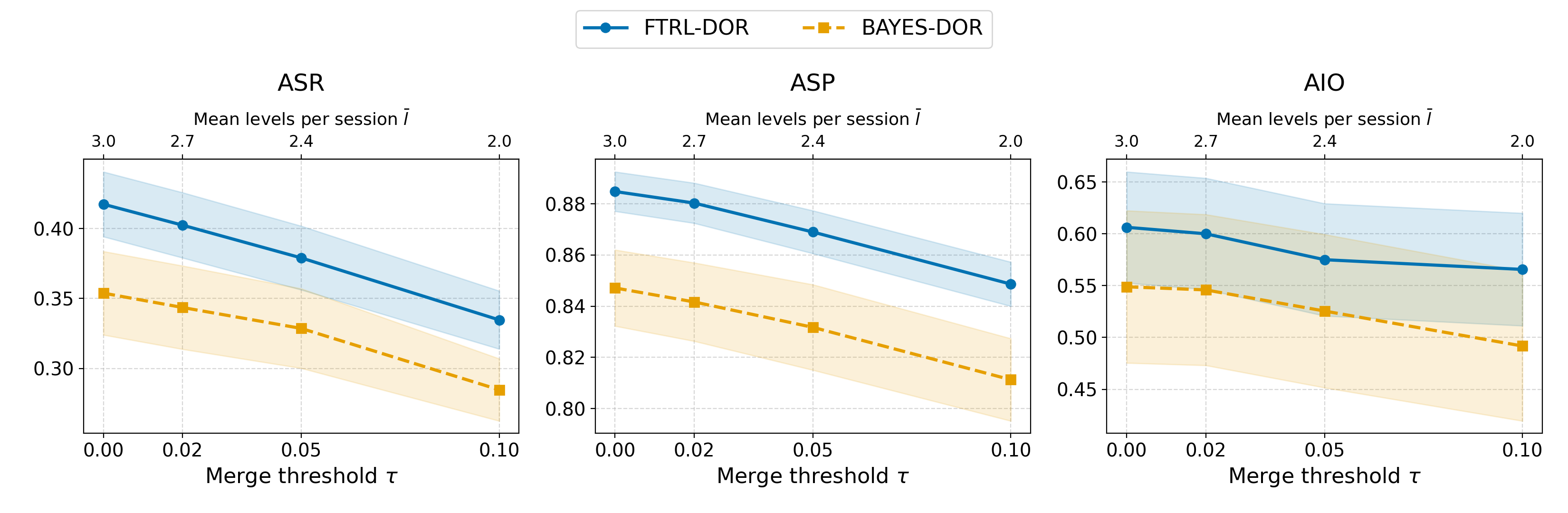}
    \caption{Ranking metric scores as a function of the merge threshold $\tau$,
    for $|A|=10$, $m=3$, $k=3$, pooled over the inconsistency levels. Top axis reports the resulting mean number of levels per session, $\bar{l}=\displaystyle\frac{1}{T}\displaystyle\sum_t l_t$.
    Shaded bands are $95\%$ confidence intervals.}
    \label{fig:sets_tau}
\end{figure}

Our framework as defined in Section~\ref{sec:b_dor} allows for ties. When the DM cannot separate two alternatives, they place them in the same level, and the level enters the likelihood through the average of its members' feature vectors, as in \eqref{eq:diff_vector}. In order to study exactly what the missing information induced by a tie costs, we introduce ties between alternatives whose true utility value distances are less than a threshold $\tau$; then we order the levels by their mean value, and generate the cards from the level means, as in Section~\ref{sec:sim_design}. We fix $|A|=10$, $m=3$, $k=3$, keep all four inconsistency levels, and let $\tau$ range over $\{0,0.02,0.05,0.10\}$, repeating experiments with same data across $\tau$ values so that the comparison is paired. At $\tau=0.10$ about $37\%$ of the sessions contain at least one tie, the average number of levels falls from $3$ to $2.03$. Figure~\ref{fig:sets_tau} shows the way ranking metrics degrade as ties are introduced. All metrics drop smoothly and moderately: ASR from $0.418$ to $0.335$ for FTRL-DOR and from $0.354$ to $0.285$ for BAYES-DOR, with AIO being the least affected metric. Ties do not degrade inference in a meaningful way. The model, in fact, pays only the information they hold back: one gap observation per merged alternative. The results show overall robustness to ties.

\subsection{Sequential prediction and regret bound}
\label{sec:sim_regret}
\begin{figure}[h!tp]
    \centering
    \includegraphics[width=0.8\textwidth]{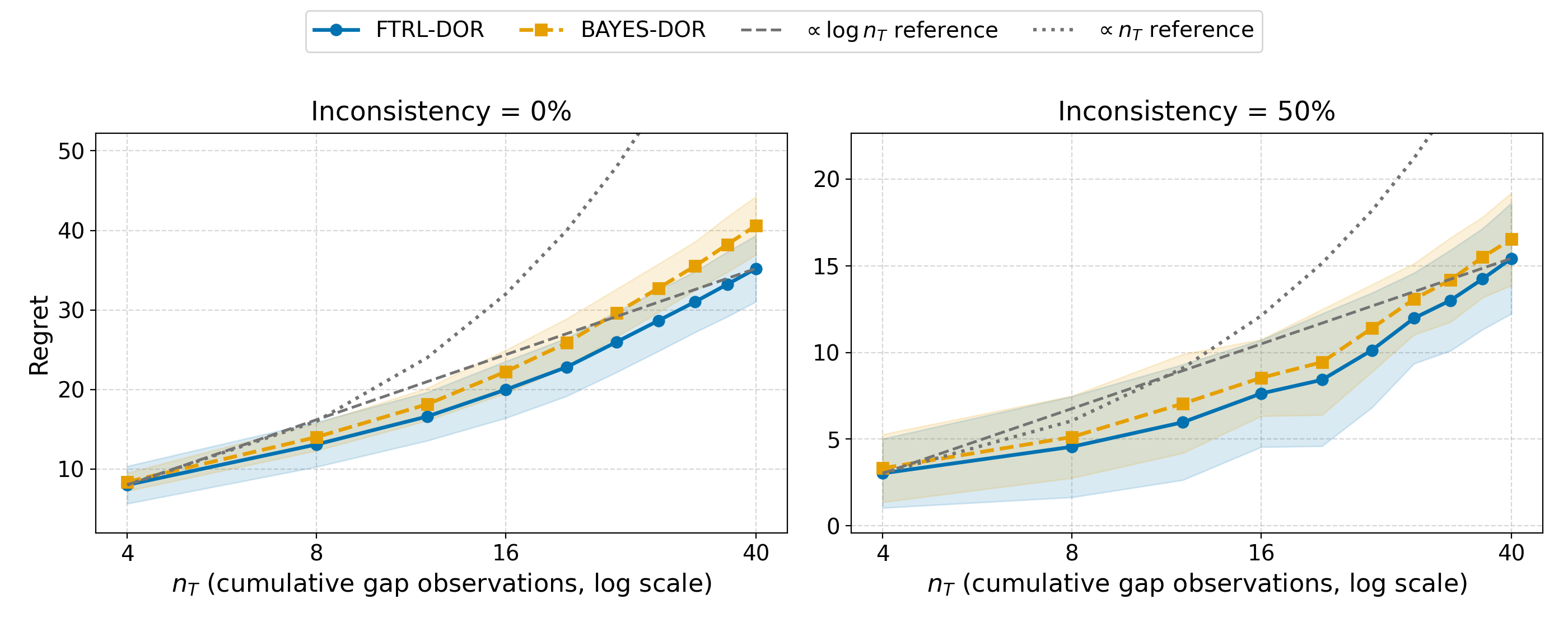}
    \caption{Regret of the two algorithms against the number of gap observations $n_T$ on a logarithmic axis, for the configuration $|A|=10$, $m=3$, $k=5$, $T=10$ (mean over five replications; shaded bands are $95\%$ confidence intervals).
    }
    \label{fig:logloss}
\end{figure}

The guarantees we provide in Section~\ref{sec:b_dor} bound the regret \eqref{eq:regret} and confer nice properties in terms of prediction of the single gap observations, but do not extend directly to ranking metrics, although we showed that robustness is empirically inherited. 
Now we verify such guarantees in the form they promise to hold. During the simulations we recorded the prequential log-loss \eqref{eq:logloss}: before each session, the algorithm assigns a probability to the DM's possible answers and we record the negative log-probability of the one actually given. We then compare it with the loss of the best fixed model in hindsight: the maximum-likelihood fit calculated using all collected gap observations produced by the end of last session. Figure~\ref{fig:logloss} shows the resulting regret for $|A|=10$, $m=3$, $k=5$, $T=10$ ($n_T=40$ gap observations), on log x-axis, so that logarithmic growth is a straight line. With a consistent DM the cumulative loss stays below the leading term of the theoretical BAYES-DOR bound ($35.2$ nats for FTRL-DOR and $40.6$ for BAYES-DOR, against $A_{\rho}\log n_T\approx52$; the FTRL-DOR coefficient involves the curvature constant of Assumption~\ref{ass:persistent_curvature} and is not directly computable); with a strongly inconsistent DM even the hindsight model loses at a linear rate, but the extra cost paid by the algorithms remains small ($15.4$ and $16.5$ nats). In both cases the regret follows the logarithmic reference: after a few sessions the algorithms predict the DM's next answers essentially as well as the best model chosen in hindsight, consistent DM or not. The comparison is conservative given the additional error factor: the recorded BAYES-DOR loss uses the finite Monte Carlo predictive rather than the exact one.

\section{An illustrative example of the B-DOR application}
\label{sec:example}
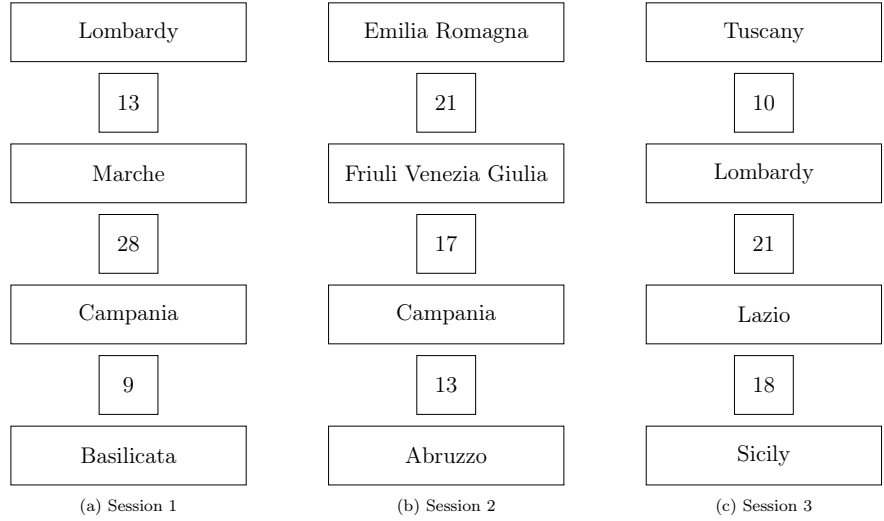
\begin{figure}[h!tp]
\centering
\resizebox{0.7\textwidth}{!}{%
\begin{subfigure}[t]{0.3\textwidth}
\centering
    \begin{tikzpicture}
\node[draw, minimum size=0.5cm, minimum width = 40mm, minimum height = 10mm] (a) at (0,0) {Lombardy};
\node[draw, minimum size=1cm, below of = a, node distance= 12mm] (b) {13};
\node[draw, minimum size=0.5cm, minimum width = 40mm, below of = b, node distance= 12mm, minimum height = 10mm] (c) {Marche};
\node[draw, minimum size=1cm, below of = c, node distance= 12mm] (d) {28};
\node[draw, minimum size=0.5cm, minimum width = 40mm, below of = d, node distance= 12mm, minimum height = 10mm] (e) {Campania};
\node[draw, minimum size=1cm, below of = e, node distance= 12mm] (f) {9};
\node[draw, minimum size=0.5cm, minimum width = 40mm, below of = f, node distance= 12mm, minimum height = 10mm] (g) {Basilicata};
\end{tikzpicture}
\subcaption{Session 1}
    \label{fig:example_t3_1}
\end{subfigure}\hfill
\begin{subfigure}[t]{0.3\textwidth}
\centering
    \begin{tikzpicture}
\node[draw, minimum size=0.5cm, minimum width = 40mm, minimum height = 10mm] (a) at (0,0) {Emilia Romagna};
\node[draw, minimum size=1cm, below of = a, node distance= 12mm] (b) {21};
\node[draw, minimum size=0.5cm, minimum width = 40mm, below of = b, node distance= 12mm, minimum height = 10mm] (c) {Friuli Venezia Giulia};
\node[draw, minimum size=1cm, below of = c, node distance= 12mm] (d) {17};
\node[draw, minimum size=0.5cm, minimum width = 40mm, below of = d, node distance= 12mm, minimum height = 10mm] (e) {Campania};
\node[draw, minimum size=1cm, below of = e, node distance= 12mm] (f) {13};
\node[draw, minimum size=0.5cm, minimum width = 40mm, below of = f, node distance= 12mm, minimum height = 10mm] (g) {Abruzzo};
\end{tikzpicture}
\subcaption{Session 2}
    \label{fig:example_t3_2}
\end{subfigure}\hfill
\begin{subfigure}[t]{0.3\textwidth}
\centering
    \begin{tikzpicture}
\node[draw, minimum size=0.5cm, minimum width = 40mm, minimum height = 10mm] (a) at (0,0) {Tuscany};
\node[draw, minimum size=1cm, below of = a, node distance= 12mm] (b) {10};
\node[draw, minimum size=0.5cm, minimum width = 40mm, below of = b, node distance= 12mm, minimum height = 10mm] (c) {Lombardy};
\node[draw, minimum size=1cm, below of = c, node distance= 12mm] (d) {21};
\node[draw, minimum size=0.5cm, minimum width = 40mm, below of = d, node distance= 12mm, minimum height = 10mm] (e) {Lazio};
\node[draw, minimum size=1cm, below of = e, node distance= 12mm] (f) {18};
\node[draw, minimum size=0.5cm, minimum width = 40mm, below of = f, node distance= 12mm, minimum height = 10mm] (g) {Sicily};
\end{tikzpicture}
\subcaption{Session 3}
    \label{fig:example_t3_3}
\end{subfigure}}
  \caption{DM preference information for each of three sessions. Each alternative populates its own level and is randomly drawn from $A^R$ in Table~\ref{tab:LEA_data}. Alternatives are ranked, and, blank cards are placed.}
    \label{fig:example_t3}
\end{figure}

In this section we show through a real-world example how the methodology proposed in Section \ref{sec:algorithm} can be used to infer alternatives' score, and hence, a ranking. In particular, let us suppose the DM wants to rank healthcare systems of Italian regions by means of the \enquote{Essential Levels of Care} (LEA), and has access to scores on $m=3$ macro criteria publicly offered by the Italian Ministry of Healthcare (Table \ref{tab:LEA_data}): $g_1$ (Prevention care); $g_2$ (Primary care); $g_3$ (Hospital care). This will be our reference set of alternatives $A^R$, with $|A^R| = 21$ number of alternatives. All criteria have increasing preference direction $(\uparrow)$ and criteria scores are bounded $g_i(a) \in [0,100]$, $\forall a \in A^R$. We set the maximum number of cards between levels $e_{max} = 30$, and we run algorithms for $T=3$ sessions. Gap observations provided at each session are summarised in Figure~\ref{fig:example_t3}. In the following subsections, we illustrate the application of the B-DOR algorithms with two different value functions.
\begin{itemize}
    \item Value function defined by a simple weighted sum across criteria $U(a) = \sum_{j=1}^m g_j(a)w_j$, obtained by setting $\gamma_j=1$ $\forall j \in \{1, \ldots, m \}$ (Section~\ref{sec:basic_concept}). In this trivial case, the parameter vector entries (weights) directly measure inferred importance of each criterion. We show concentration of weights over sessions (Section \ref{sec:example_linear}).
    \item Value function defined as the sum of piecewise marginal value functions, each with four characteristic points per criterion: $\gamma_j=3$, $\forall j \in \{1, \ldots, m \}$. This is the general case that allows inference of non-linear DM response to criteria scores. We show evolution of marginal value functions over sessions (Section \ref{sec:example_pw}).
\end{itemize} 
In both cases, we produce rankings using FTRL-DOR and BAYES-DOR algorithms. Specifically we run Algorithm~\ref{eq:algo} with the input parameters we just defined. FTRL-DOR outputs, at each session $t$, one parameter vector $\Phi^{\star}_t = (\boldsymbol{w}^{\star}_t, \boldsymbol{\Delta}^{\star}_t)$, where $\boldsymbol{\Delta}^{\star}$ parametrises a DM function that maps a real valued pairwise preference strength to an integer number of cards, and $\boldsymbol{w}^{\star}$ parametrises the value function, hence the ranking. BAYES-DOR outputs a set $\Phi_t^R = \{\boldsymbol{\phi}_t^1, \ldots, \boldsymbol{\phi}_t^R\}$ of samples from the posterior. In both algorithms $\boldsymbol{w} \in \mathbb{R}^N_+$ so $u(\cdot)_j \in \mathbb{R}_+$ (Section~\ref{sec:basic_concept}). Marginals require normalization to gain ranking interpretability. The obvious normalization choice, with bounded criteria scores, is dividing by the maximum global score: $\hat{u}_j(\cdot) = \displaystyle\frac{u_j(\cdot)}{U^{\star}}$, where $U^{\star} = \displaystyle\sum_{j=1}^{m} u_j(100)$. Having a maximum common value, and being their sum fixed, normalised inferred marginals $\hat{u}_j$ have direct association to the ranking and can be interpreted by the DM. The prior parameters are set to $\alpha=\alpha_{\Delta}=1,\beta=\beta_{\Delta}=1$ for BAYES-DOR and $\alpha=\alpha_{\Delta}=2,\beta=\beta_{\Delta}=1$ for FTRL-DOR. Again, the choice of these values is implied by the regret theory, as discussed in Section~\ref{sec:hyper}. We fix the number of samples generated by BAYES-DOR to $R=2\cdot10^3$, with $2\cdot10^3$ discarded warm-up samples, and check their Monte Carlo convergence statistic throughout the procedure in the same way as described in Section~\ref{sec:sim_design}.
\begin{table}[h!]
    \centering
    \footnotesize
    \caption{Italian regions' performance on the LEA criteria in 2024
    (\href{https://www.salute.gov.it/new/it/tema/livelli-essenziali-di-assistenza/il-nuovo-sistema-di-garanzia-nsg/?paragraph=0}{source}).}
    \begin{tabular}{llccc}
    \hline
    ID & Region & Prevention care - $g_1$ ($\uparrow$) & Primary care - $g_2$ ($\uparrow$) & Hospital care - $g_3$ ($\uparrow$) \\
    \hline
    $R_1$ & Piedmont & 93 & 90 & 87 \\
    $R_2$ & Aosta Valley & 77 & 35 & 53 \\
    $R_3$ & Lombardy & 95 & 76 & 86 \\
    $R_4$ & Autonomous Province of Bolzano & 58 & 82 & 62 \\
    $R_5$ & Autonomous Province of Trento & 98 & 83 & 97 \\
    $R_6$ & Veneto & 98 & 96 & 94 \\
    $R_7$ & Friuli Venezia Giulia & 81 & 81 & 73 \\
    $R_8$ & Liguria & 54 & 85 & 80 \\
    $R_9$ & Emilia-Romagna & 97 & 89 & 92 \\
    $R_{10}$ & Tuscany & 95 & 95 & 96 \\
    $R_{11}$ & Umbria & 93 & 80 & 84 \\
    $R_{12}$ & Marche & 74 & 83 & 91 \\
    $R_{13}$ & Lazio & 63 & 68 & 85 \\
    $R_{14}$ & Abruzzo & 54 & 45 & 83 \\
    $R_{15}$ & Molise & 58 & 73 & 62 \\
    $R_{16}$ & Campania & 62 & 72 & 72 \\
    $R_{17}$ & Apulia & 74 & 69 & 85 \\
    $R_{18}$ & Basilicata & 68 & 52 & 69 \\
    $R_{19}$ & Calabria & 68 & 40 & 69 \\
    $R_{20}$ & Sicily & 49 & 44 & 80 \\
    $R_{21}$ & Sardinia & 65 & 67 & 60 \\
    \hline
    \end{tabular}
    \label{tab:LEA_data}
\end{table}

\subsection{Linear marginal value functions}
\label{sec:example_linear}
\begin{figure}[h!]
    \centering
\includegraphics[trim = 0 0 0 0, clip,width=0.5\textwidth]{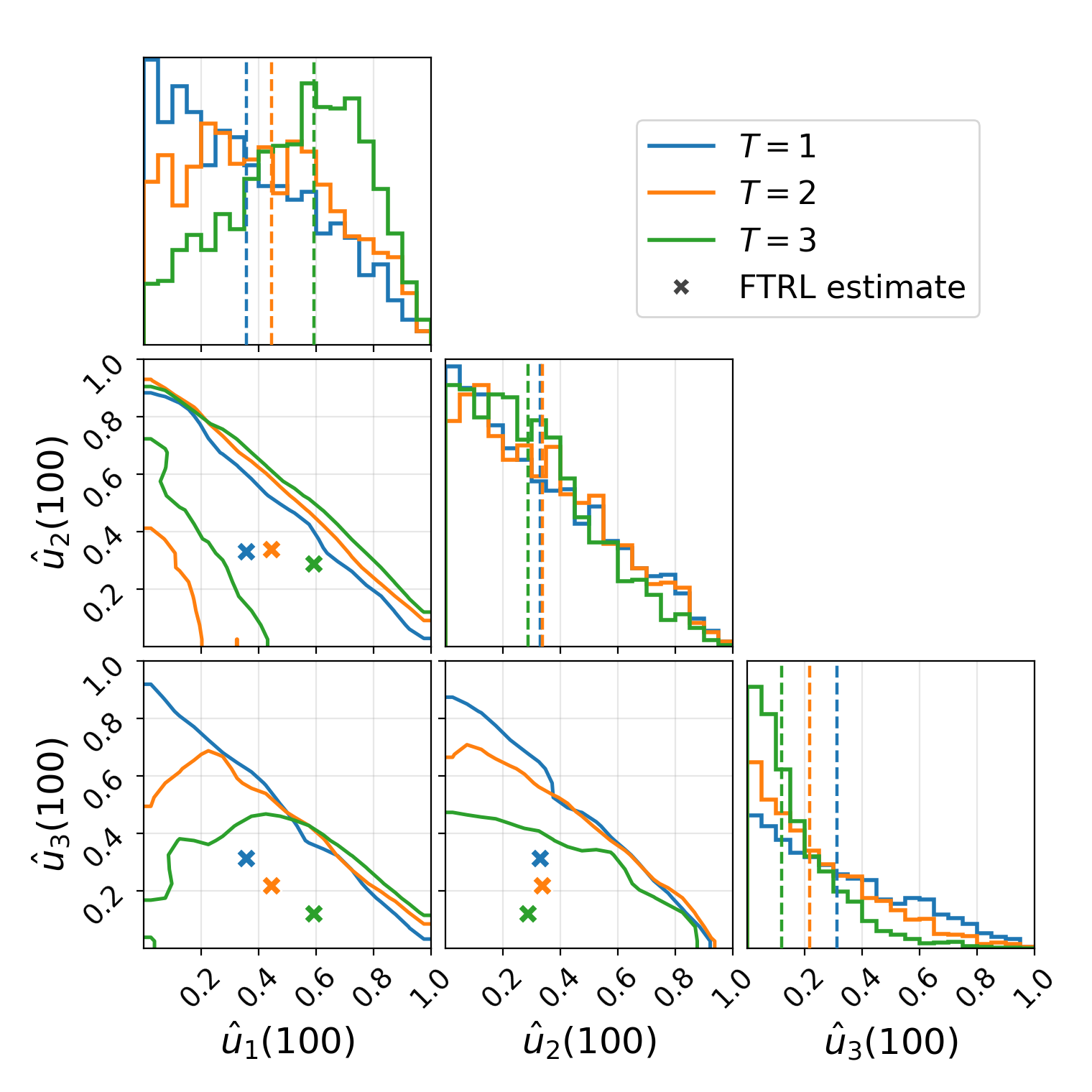}
    \caption{Corner plot of the posterior distribution of the normalised marginal value functions $\left[\hat{u}_1(100), \hat{u}_2(100), \hat{u}_3(100)\right]$ after each of the $T=3$ sessions ($T=1$ in blue, $T=2$ in orange, $T=3$ in green). 
    Diagonal panels show the marginal distributions with the FTRL-DOR estimates (dashed vertical lines); off-diagonal panels show the $90\%$ credible contours of the bivariate distributions with the FTRL-DOR estimates (crosses).}
    \label{fig:croner_lin}
\end{figure}

We first assume linear marginal value functions, that is, two characteristic points per criterion. With this choice, each marginal value function is a straight line, so the model has a single parameter per criterion and $\hat{u}_j(100)=\displaystyle\frac{100\cdot w_j}{U^{\star}}$ is the maximum normalised contribution of the marginal evaluation of criterion $j$ to the global evaluation. Figure~\ref{fig:croner_lin} shows the corner plot of the posterior distribution of $\left(\hat{u}_1(100), \hat{u}_2(100), \hat{u}_3(100)\right)$ estimated from samples $\{w_j^1, \ldots, w_j^R\}$ collected from BAYES-DOR. A different set of samples is collected after each session ($T=1$ in blue, $T=2$ in orange, $T=3$ in green), together with the FTRL-DOR estimate (dashed vertical lines on the diagonal panels, crosses on the bivariate panels).

The sessions act on the posterior in two distinct ways. First, they shift it: the evidence progressively favors Prevention care ($g_1$), whose median share grows from $0.32$ after the first session to $0.56$ after the third, at the expense of Hospital care ($g_3$), which falls from $0.27$ to $0.12$. The FTRL-DOR estimates follow the same path ($0.36 \to 0.45 \to 0.59$ for $g_1$ and $0.31 \to 0.22 \to 0.12$ for $g_3$), and they are close to the posterior medians at every session. Second, the posterior concentrates across sessions, but not to the same extent on the three criteria. The $90\%$ credible interval of $\hat{u}_3(100)$ (Hospital care) is the one that narrows most, from $[0.02, 0.76]$ to $[0.01, 0.46]$; that of $\hat{u}_2(100)$ (Primary care) narrows moderately, from $[0.03, 0.78]$ to $[0.03, 0.70]$; while that of $\hat{u}_1(100)$ (Prevention care) keeps almost the same width, moving from $[0.03, 0.81]$ to $[0.11, 0.88]$. What changes for Prevention care is the lower end of the interval, which rises from $0.03$ to $0.11$: the sessions exclude that this criterion is negligible. This suggests that the DM is consistent with the order of preference of criteria: Prevention care first, then Primary care, and Hospital care last.

Table~\ref{tab:rank_lin_top5} reports the five highest-scoring regions after each session according to FTRL-DOR, along with their inferred normalised overall evaluations $\hat{U}(a) =\displaystyle \sum_{j=1}^m \hat{u}_j(g_j(a))$. Both composition and order of these top-five alternatives do not change over sessions: Veneto, Tuscany, Emilia Romagna, Autonomous Province of Trento, Piedmont; but their margins do: the gap between Veneto and Tuscany widens from $0.8$ points after the first session to $1.8$ after the third, while Emilia Romagna and Trento stay within about half a point of each other at every session. Margins this small raise the question of how much confidence the DM should place in the estimated order, a question that the point estimates cannot answer by themselves.

\begin{table}[h!]
\centering
\small
\caption{The five highest-scoring regions after each of the $T=3$ sessions according to FTRL-DOR, with their normalised inferred evaluations. Alternatives included in the DM preferences are written in bold.}
\label{tab:rank_lin_top5}
\begin{tabularx}{\textwidth}{c|Cc|Cc|Cc}
\toprule
\multirow{2}{*}{Rank}&\multicolumn{2}{c}{$T=1$}&\multicolumn{2}{c}{$T=2$}&\multicolumn{2}{c}{$T=3$}\\
\cmidrule(lr){2-3}\cmidrule(lr){4-5}\cmidrule(lr){6-7}
&Region&$\hat{U}\left(a\right)$&Region&$\hat{U}\left(a\right)$&Region&$\hat{U}\left(a\right)$\\
\midrule
1 & Veneto & 96.09 & Veneto & 96.46 & Veneto & 96.94 \\
2 & Tuscany & 95.31 & Tuscany & 95.22 & \textbf{Tuscany} & \textbf{95.12} \\
3 & Emilia Romagna & 92.80 & \textbf{Emilia Romagna} & \textbf{93.22} & \textbf{Emilia Romagna} & \textbf{94.10} \\
4 & A. P. of Trento & 92.74 & A. P. of Trento & 92.72 & A. P. of Trento & 93.58 \\
5 & Piedmont & 90.13 & Piedmont & 90.69 & Piedmont & 91.41 \\
\bottomrule
\end{tabularx}
\end{table}

The RAI \eqref{eq:RAI} and the PWI \eqref{eq:PWI}, computed from the posterior sample of each session, answer precisely this question. Table~\ref{tab:rai_evo_lin} reports the RAI of the four regions that occupy more often the first four positions of the ranking after each session, and Table~\ref{tab:pwi_evo_lin} the PWI among them.

First thing we notice is that Veneto is ranked first for most weight vectors already after the single first session: its rank-1 acceptability is $67.7\%$ at $T=1$. As the posterior then drifts toward Prevention care (the criterion where Veneto's advantage over its closest competitor Tuscany is largest) and away from Hospital care (the only criterion where Tuscany is ahead) the rank-1 acceptability climbs to $79.3\%$ and $88.5\%$, and the PWI of Veneto over Tuscany from $76.3\%$ to $95.8\%$. After the third session Veneto's PWI against every other region is at least $91.5\%$. FTRL-DOR tells the same story in point-estimate form, ranking Veneto first after every session.

Rank-samples are useful to identify and quantify uncertainty that involves alternatives for which the inferred global evaluations are close. Emilia Romagna and the Autonomous Province of Trento occupy the third and fourth positions, but their comparison is uncertain, and DM's answers constrain uncertainty only marginally across sessions. Emilia Romagna is stronger on Primary care ($89$ against $83$), Trento on Hospital care ($97$ against $92$). Their PWI stays close to one half at every session ($47.6\%$, $52.3\%$, $56.5\%$), drifting slowly toward Emilia Romagna as the Hospital care weight drops over sessions, and the rank distributions of the two regions keep overlapping on ranks $2$-$4$. To summarize: FTRL-DOR places Emilia Romagna third and Trento fourth after every session; the PWI over samples contributes with information that this ordering is supported by barely more than half of the posterior mass. This is uncertainty information that a point estimate alone cannot cover.

Moreover, the acceptability indices expose structure that a single ranking hides. For example, Tuscany's rank-1 RAI drops across sessions ($16.8\% \to 9.8\% \to 3.2\%$), while roughly two thirds of RAI score are assigned to the second rank constantly throughout sessions, meaning that the residual mass moves to the fourth rank ($10.1\% \to 13.2\% \to 21.6\%$), which corresponds to the posterior draws in which both Emilia Romagna and Trento overtake it. Tuscany is thus confirmed as the main challenger of Veneto, but with a growing, quantified possibility that it actually sits behind the Emilia Romagna and Trento pair.

\begin{table}[h!]
\centering
\footnotesize
\caption{Evolution across the $T=3$ sessions of the Rank Acceptability Indices (\%) of the four top-ranked regions, in the case of linear marginal value functions}
\label{tab:rai_evo_lin}
\begin{tabularx}{\textwidth}{l l|CCCCCCCCC}
\toprule
\multirow{2}{*}{Region}&\multirow{2}{*}{Sessions}&\multicolumn{9}{c}{Rank} \\
\cmidrule(lr){3-11}
&&1&2&3&4&5&6&7&8&$>8$\\
\midrule
\multirow{3}{*}{Veneto ($R_{6}$)}&$T=1$&67.7&22.4&9.9&0&0&0&0&0&0\\
&$T=2$&79.3&15.4&5.2&0&0&0&0&0&0\\
&$T=3$&88.5&10.2&1.2&0&0&0&0&0&0\\
\midrule
\multirow{3}{*}{Tuscany ($R_{10}$)}&$T=1$&16.8&61.2&11.9&10.1&0&0&0&0&0\\
&$T=2$&9.8&64.3&12.7&13.2&0&0&0&0&0\\
&$T=3$&3.2&63.5&11.7&21.6&0&0&0&0&0\\
\midrule
\multirow{3}{*}{Emilia Romagna ($R_{9}$)}&$T=1$&0&0.9&52.8&46.3&0&0&0&0&0\\
&$T=2$&0&1.8&59.9&38.4&0&0&0&0&0\\
&$T=3$&0&3.6&70.8&25.6&0&0&0&0&0\\
\midrule
\multirow{3}{*}{A. P. of Trento ($R_{5}$)}&$T=1$&15.6&15.4&22.1&22.9&23.6&0.4&0&0&0\\
&$T=2$&10.9&18.5&19.1&25.9&25.3&0.4&0&0&0\\
&$T=3$&8.2&22.7&14.3&33.8&20.8&0.1&0&0&0\\
\bottomrule
\end{tabularx}
\end{table}

\begin{table}[h!]
\centering
\footnotesize
\caption{Evolution across the $T=3$ sessions of the Pairwise Winning Indices (\%) among the four top-ranked regions, in the case of linear marginal value functions}
\label{tab:pwi_evo_lin}
\begin{tabularx}{\textwidth}{l|CCCC|CCCC|CCCC}
\toprule
&\multicolumn{4}{c}{$T=1$}&\multicolumn{4}{c}{$T=2$}&\multicolumn{4}{c}{$T=3$}\\
\cmidrule(lr){2-5}\cmidrule(lr){6-9}\cmidrule(lr){10-13}
Region&$R_{6}$&$R_{10}$&$R_{9}$&$R_{5}$&$R_{6}$&$R_{10}$&$R_{9}$&$R_{5}$&$R_{6}$&$R_{10}$&$R_{9}$&$R_{5}$\\
\midrule
Veneto ($R_{6}$)&--&76.3&100&81.5&--&86.4&100&87.7&--&95.8&100&91.5\\
Tuscany ($R_{10}$)&23.7&--&89.7&71.3&13.6&--&85.9&71.2&4.2&--&76.5&67.5\\
Emilia Romagna ($R_{9}$)&0&10.3&--&47.6&0&14.1&--&52.3&0&23.4&--&56.5\\
A. P. of Trento ($R_{5}$)&18.6&28.7&52.3&--&12.3&28.8&47.6&--&8.5&32.6&43.5&--\\
\bottomrule
\end{tabularx}
\end{table}

The complete RAI and PWI of all the 21 regions, computed both after the first session and after the third one, are reported in Tables \ref{tab:RAI_lin_t1}-\ref{tab:PWI_lin_t3} of \ref{appendix:rai-pwi} in the online supplementary material.

\subsection{Piecewise linear marginal value functions}
\label{sec:example_pw}
\begin{figure}[h!]
    \centering
\includegraphics[trim = 0 0 0 0, clip,width=0.8\textwidth]{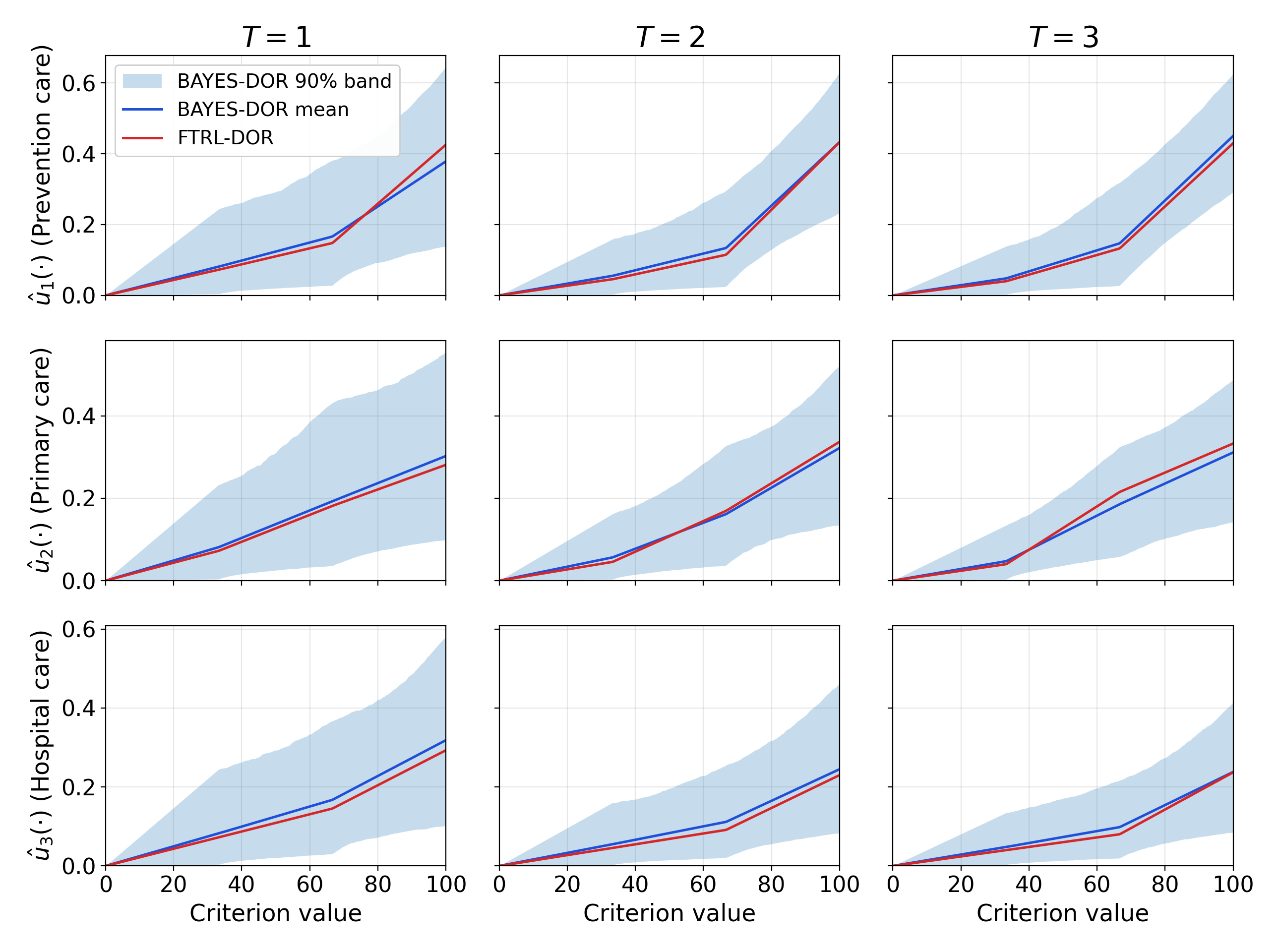}
    \caption{Posterior distribution of the normalised marginal value functions in the case of piecewise linear marginal value functions ($\gamma_j=3$): rows correspond to criteria, columns to sessions. Each panel shows the BAYES-DOR posterior mean with its $90\%$ credible band and the FTRL-DOR estimate.
    }
    \label{fig:marginals_pw}
\end{figure}

We now repeat the process using piecewise linear marginal value functions with $\gamma_j = 3$, meaning four equally spaced characteristic points per criterion, so that each marginal evaluation is described by three parameters and can assume either convex or concave shape. Figure~\ref{fig:marginals_pw} shows the resulting estimates: each row corresponds to a criterion, each column to a session, and each panel reports the BAYES-DOR posterior mean of the normalised marginal value function with its $90\%$ credible band, together with the FTRL-DOR estimate. The normalization is the same of Section~\ref{sec:example_linear}: every posterior draw is divided by the overall evaluation of the ideal alternative, so the value reached at $100$ is the importance share of the criterion.

First thing we notice is that estimated marginal value functions are clearly nonlinear, and the nonlinearity is criterion-specific: after the third session, Prevention care and Hospital care gain about two thirds of their value in the last third of the score domain (the FTRL-DOR estimate of $\hat{u}_1$ rises from $0.13$ at $66.7$ to $0.43$ at $100$, that of $\hat{u}_3$ from $0.08$ to $0.24$), while Primary care gains more than half of its value in the central third. The values at the second characteristic point (that is the first non-zero one) are learned from the prior rather than from the data: both algorithms assign $\hat{u}_j(33.3)$ small and essentially equal values across the three criteria. This is because every region scores at least $35$ on every criterion, so the first segment contributes identically to all the alternatives and cancels from every pairwise comparison in the likelihood. Rankings, RAI and PWI are essentially unaffected by the criteria evaluation before the second characteristic point. Information over sessions accumulates in a way that moves the mean estimates and reduces their uncertainty. For example Hospital care's curve flattens: $\hat{u}_3(100)$ value drops from $0.31$ to $0.23$; the bands narrow mostly in the upper part of the score domain, where the information actually lies: the $90\%$ interval of $\hat{u}_3(66.7)$ shrinks from $[0.03, 0.37]$ to $[0.02, 0.22]$ and that of $\hat{u}_1(100)$ from $[0.14, 0.65]$ to $[0.29, 0.63]$, with the FTRL-DOR curves staying inside the bands and close to the posterior means throughout ($0.430/0.333/0.237$ against posterior medians $0.442/0.308/0.233$ at $T=3$).

The comparison with Section~\ref{sec:example_linear} is instructive. The linear marginals could represent gained information over sessions only by moving their slope, pushing the Prevention care weight up to $0.59$ while leaving it poorly identified ($90\%$ credible interval $[0.11, 0.88]$ after the third session); the piecewise specification absorbs the same information through the shape of the marginals, and the shares stay more spread ($0.43/0.33/0.24$) and considerably better identified ($[0.29, 0.63]$ for Prevention care). Allowing for more complex marginal value functions can thus change the conclusions that the DM obtains from the algorithm about the importance of criteria.

We now explore the output in terms of ranking.
Table~\ref{tab:rank_pw_top5} reports the first five ranked positions by FTRL-DOR. The group is composed of the same five regions that were ranked first in the case with linear marginals, and Veneto's margin over Tuscany is even wider here. A difference from the linear case can be spotted when looking at the ranking of the third and fourth alternatives: the Autonomous Province of Trento and Emilia Romagna appear in the reverse order with respect to the linear case.

\begin{table}[h!]
\centering
\small
\caption{The five highest-scoring regions after each of the $T=3$ sessions according to FTRL-DOR, with their normalised inferred piecewise linear marginal value functions ($\gamma_j=3$).
Alternatives included in the DM preferences are written in bold.}
\label{tab:rank_pw_top5}
\begin{tabularx}{\textwidth}{c|Cc|Cc|Cc}
\toprule
\multirow{2}{*}{Rank}&\multicolumn{2}{c}{$T=1$}&\multicolumn{2}{c}{$T=2$}&\multicolumn{2}{c}{$T=3$}\\
\cmidrule(lr){2-3}\cmidrule(lr){4-5}\cmidrule(lr){6-7}
&Region&$U\left(a\right)$&Region&$U\left(a\right)$&Region&$U\left(a\right)$\\
\midrule
1 & Veneto & 94.48 & Veneto & 93.59 & Veneto & 93.98 \\
2 & Tuscany & 92.57 & Tuscany & 91.05 & \textbf{Tuscany} & \textbf{91.90} \\
3 & A. P. of Trento & 91.92 & A. P. of Trento & 88.30 & A. P. of Trento & 90.81 \\
4 & Emilia Romagna & 90.67 & \textbf{Emilia Romagna} & \textbf{88.28} & \textbf{Emilia Romagna} & \textbf{89.68} \\
5 & Piedmont & 85.42 & Piedmont & 82.89 & Piedmont & 84.12 \\
\bottomrule
\end{tabularx}
\end{table}

Tables~\ref{tab:rai_evo_pw} and \ref{tab:pwi_evo_pw} show the evolution of the RAI and PWI for this leading group. The ranked position of Veneto is confirmed and even amplified: its rank-1 acceptability grows from $65.9\%$ after the first session to $83.0\%$ after the third, and its PWI over Tuscany reaches $98.2\%$. The comparison between Trento and Emilia Romagna remains a weak point of the recommendation: 
under linear marginal value functions Emilia Romagna ends slightly ahead (PWI $56.5\%$), under piecewise value functions Trento does (PWI $63.6\%$, with a rank-1 acceptability of $16.3\%$, twice the $8.2\%$ of the linear case). Notably, the choice of the value function itself can tip the balance.

\begin{table}[h!]
\centering
\footnotesize
\caption{Evolution across the $T=3$ sessions of the Rank Acceptability Indices (\%) of the four top-ranked regions, in the case of piecewise linear marginal value functions ($\gamma_j=3$)}
\label{tab:rai_evo_pw}
\begin{tabularx}{\textwidth}{l l|CCCCCCCCC}
\toprule
\multirow{2}{*}{Region}&\multirow{2}{*}{Sessions}&\multicolumn{9}{c}{Rank} \\
\cmidrule(lr){3-11}
&&1&2&3&4&5&6&7&8&$>8$\\
\midrule
\multirow{3}{*}{Veneto ($R_{6}$)}&$T=1$&65.9&23.8&10.3&0&0&0&0&0&0\\
&$T=2$&84.2&12.8&2.9&0&0&0&0&0&0\\
&$T=3$&83.0&15.8&1.2&0&0&0&0&0&0\\
\midrule
\multirow{3}{*}{Tuscany ($R_{10}$)}&$T=1$&13.3&51.9&19.8&15.0&0&0&0&0&0\\
&$T=2$&2.9&65.9&19.7&11.6&0&0&0&0&0\\
&$T=3$&0.8&59.9&30.5&8.9&0&0&0&0&0\\
\midrule
\multirow{3}{*}{A. P. of Trento ($R_{5}$)}&$T=1$&20.8&22.4&23.8&20.4&12.5&0&0&0&0\\
&$T=2$&12.9&19.8&21.8&33.4&12.2&0.1&0&0&0\\
&$T=3$&16.3&22.9&25.3&29.4&6.0&0&0&0&0\\
\midrule
\multirow{3}{*}{Emilia Romagna ($R_{9}$)}&$T=1$&0&1.8&45.4&52.8&0&0&0&0&0\\
&$T=2$&0&1.6&55.2&43.2&0&0&0&0&0\\
&$T=3$&0&1.4&43.0&55.6&0&0&0&0&0\\
\bottomrule
\end{tabularx}
\end{table}

\begin{table}[h!]
\centering
\footnotesize
\caption{Evolution across the $T=3$ sessions of the Pairwise Winning Indices (\%) among the four top-ranked regions, in the case of piecewise linear marginal value functions ($\gamma_j=3$)}
\label{tab:pwi_evo_pw}
\begin{tabularx}{\textwidth}{l|CCCC|CCCC|CCCC}
\toprule
&\multicolumn{4}{c}{$T=1$}&\multicolumn{4}{c}{$T=2$}&\multicolumn{4}{c}{$T=3$}\\
\cmidrule(lr){2-5}\cmidrule(lr){6-9}\cmidrule(lr){10-13}
Region&$R_{6}$&$R_{10}$&$R_{5}$&$R_{9}$&$R_{6}$&$R_{10}$&$R_{5}$&$R_{9}$&$R_{6}$&$R_{10}$&$R_{5}$&$R_{9}$\\
\midrule
Veneto ($R_{6}$)&--&79.3&76.2&100&--&94.8&86.6&100&--&98.2&83.5&100\\
Tuscany ($R_{10}$)&20.6&--&58.7&84.2&5.2&--&67.2&87.6&1.8&--&60.1&90.6\\
A. P. of Trento ($R_{5}$)&23.8&41.3&--&66.0&13.5&32.9&--&53.6&16.5&40.0&--&63.6\\
Emilia Romagna ($R_{9}$)&0&15.8&34.1&--&0&12.3&46.4&--&0&9.3&36.4&--\\
\bottomrule
\end{tabularx}
\end{table}

The complete RAI and PWI of all the 21 regions, computed both after the first session and after the third one, are reported in Tables \ref{tab:RAI_pw_t1}-\ref{tab:PWI_pw_t3} of \ref{appendix:rai-pwi} in the online supplementary material.

\section{Conclusion}
\label{sec:conclusion}
The main contribution of this paper is to bind the Deck-of-cards-based Ordinal Regression with a probabilistic Bayesian framework that we call Bayesian DOR (B-DOR). The Decision Maker (DM) preference information (the rank-order and blank cards) is treated as random observation rather than being used to build constraints that a value function must satisfy. On this model, B-DOR builds two online inference algorithms: BAYES-DOR, which maintains an entire posterior distribution over value functions and allows it to provide robust recommendations such as RAI and PWI; FTRL-DOR, which outputs the value function that maximizes the posterior and is found by constrained convex optimization. The model is designed to be a multi-session elicitation process where the posterior of one session is the prior of the next, and the elicitation can be spread over several short sessions that may reduce the DM cognitive burden.

On the theoretical side, both algorithms come with regret guarantees (Theorems~\ref{thm:bayes_dor_regret} and~\ref{thm:ftrl_dor_regret}): their cumulative prediction loss exceeds that of the best value function chosen in hindsight only by a term logarithmic in the number of gap observations, for every realised sequence of answers and without stochastic assumptions on the DM (for FTRL-DOR, under a curvature condition on the collected observations). The bounds concern the prediction of the DM's answers rather than the recovery of the true ranking;
Section~\ref{sec:sim_regret} verifies the logarithmic growth in practice, and the simulation study shows that the promised robustness is inherited by the ranking metrics.

A simulation study based on 768 configurations supports four main conclusions: (i) Performance metrics improve with every session, and having more small sessions rather than few larger ones is the right choice for an inconsistent DM; (ii) The blank cards are worth eliciting: both algorithms significantly outperform their direction-only counterparts on every metric, and a higher card resolution (higher $e_{max}$) should be preferred when the DM inconsistency is estimated to grow sub-linearly with $e_{max}$. (iii) Both algorithms exploit strong robustness to inconsistent answers. 
(iv) Even when the algorithms undergo a single-session, the case for which DOR and SMAA-DOR are designed, FTRL-DOR outperforms DOR and BAYES-DOR outperforms SMAA-DOR on every considered metric.

We show the potential of the model with a didactic example based on a real-world case study where Italian regions are evaluated on their Healthcare system based on three criteria: Prevention care, Primary care and Hospital care. Beyond this example, the proposed methodology has the potential to construct composite indicators in any domain.

Several topics are worth future discussion. First, all reported evidence rests on preferences generated by a single synthetic noise model: a behavioural study with real DMs is needed to validate the card-response model. Second, the choice of the subset presented at each session is left to the analyst: active learning strategies that select the queries with maximal expected information gain \citep{ciomek2017heuristics}, paired with stopping rules based on the posterior spread, would shorten the elicitation further. Finally, more complex frameworks can be built to allow the DM to place an interval of number of cards \citep{corrente2017robust}; use preference models beyond the additive one, such as the Choquet integral \citep{choquet1954theory}; include a hierarchical structure of criteria \citep{corrente2012multiple}; apply the methods to sorting problems. \citep{devaud1980utadis}.

\section*{Declaration of generative AI and AI-assisted technologies in the manuscript preparation process}
During the preparation of this work the authors used Claude (Anthropic) in order to review the flow writing, and to generate the Python codes. After using this tool, the authors reviewed and edited the content as needed and take full responsibility for the content of the published article.
\section*{Acknowledgements}
Marco Grillo acknowledge support from the HORIZON-MSCA-2021-DN-01 project LEMUR no. 101073307, funded by the European Union.

\bibliographystyle{agsm}
\bibliography{Full_bibliography}

\end{document}